# Computational Machines in a Coexistence with Concrete Universals and Data Streams

Vahid Moosavi
Chair for Computer Aided Architectural Design, Department of Architecture, ETH Zurich, 8093, Switzerland

## Abstract

We discuss that how the majority of traditional modeling approaches are following the idealism point of view in scientific modeling, which follow the set theoretical notions of models based on abstract universals. We show that while successful in many classical modeling domains, there are fundamental limits to the application of set theoretical models in dealing with complex systems with many potential aspects or properties depending on the perspectives. As an alternative to abstract universals, we propose a conceptual modeling framework based on concrete universals that can be interpreted as a category theoretical approach to modeling. We call this modeling framework pre-specific modeling.

We further, discuss how a certain group of mathematical and computational methods, along with ever-growing data streams are able to operationalize the concept of pre-specific modeling.

## 1. How to Approach the Notion of Scientific Modeling?

Modeling paradigms, as a necessary element of any scientific investigation, act like pairs of glasses, which impact the way in which we encode (conceive of) the real world. Therefore, any kind of intervention in real world phenomena is affected by the chosen modeling paradigm and the real phenomena under investigation. In the domain of urbanism and urban design, cities as complex and open environments with dynamic and multidimensional aspects are challenging cases for modeling scholars, as there are many distinct urban phenomena. Figure 1 shows a list of different functional aspects of urban phenomena in an indexical manner.

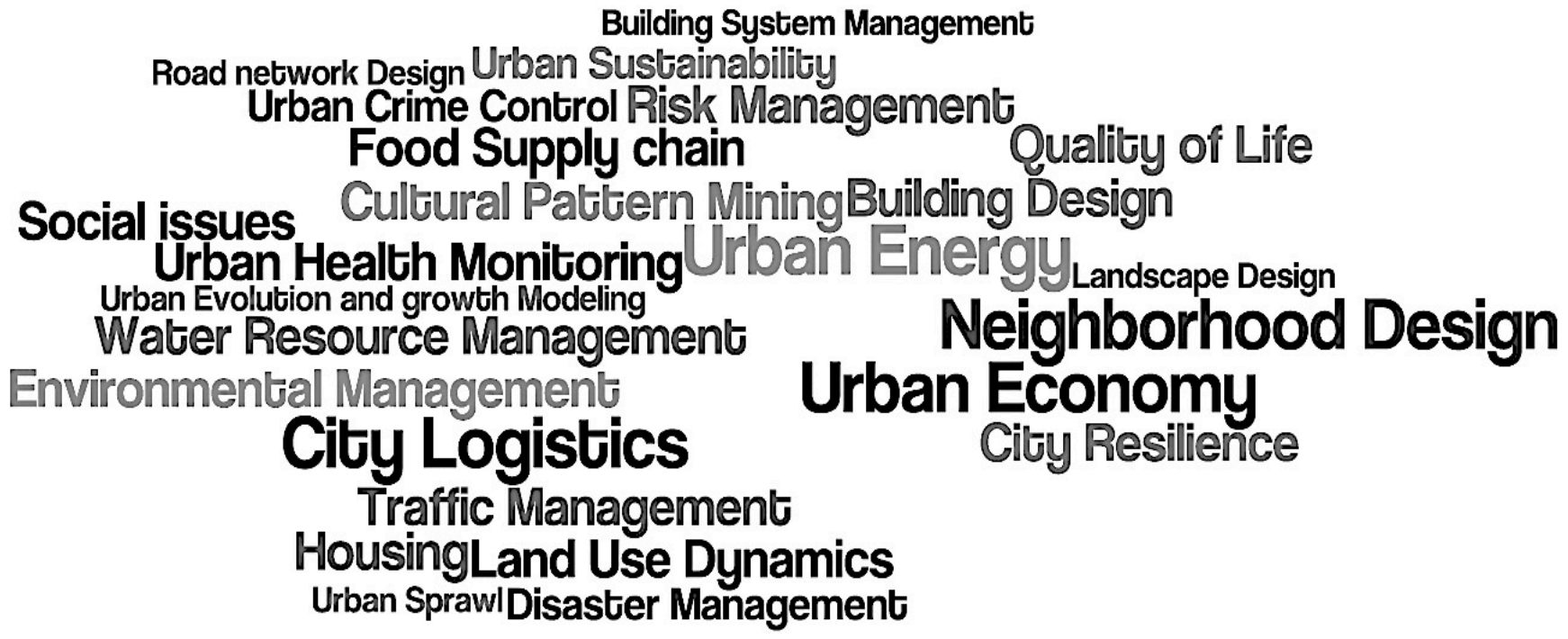

Figure 1. Different functional aspects of urban phenomena

In addition to the diversity of urban problems, there is a huge variety of competing paradigms for analyzing cities: the city as an ecological phenomena that is optimally adjusted to an environment (economic, political, cultural) assumed to be “natural” for it; the city as a thermodynamic system that needs to be balanced and which can be controlled; the city as a grammatical text with its own “syntactical laws”; the city as a biological organism following fractal growth patterns. Further, historical perspective provide additional city models such as the City of Faith, the City as Machine, or the Organic City,[1] and especially, since the advent of computers from the second half of the twentieth century, city as information.[2]

Although comparing to classical science and its engineering disciplines such as Physics, Chemistry and Mechanics, urban design, planning and modeling is a rather young discipline, when one does a quick search of the keywords central to this field, one is quickly confused by the number of approaches and the variety of practical problems within the reaches of the discipline. For example, A. G. Wilson's five volume text on urban modeling is over 2,600 pages long.[3] A broad range of case-based canonization has thus emerged, and applied techniques are developed for specific urban functions such as urban land use, urban transportation, urban economy, urban social patterns, and so on. As a result, the lack of a more abstract categorization of applied techniques makes comparison between them very hard.

Beginning in mid-twentieth century, General Systems Theory emerged as one of the main theories for working toward unification of different disciplinary modeling practices.[4] In principle, the underlying idea of systems theory is the promotion of a unified view to modernist-reductionist science, which was diversified around a variety of application and functional domains. Although, interdisciplinary collaborations such as making analogies within disciplines, (e.g., hydraulic theories to describe biological systems) was not new, systems theory's formalization, as an orthogonal view to classically diversified scientific and practical problems, reached to a point in which, according to George Klir, systemic tasks such as modeling, optimization, and simulation have emerged as distinct scientific disciplines.[5] However, taking systems theory as a body of knowledge (rather than a specific and singular theory), one could expect a gradual divergence of its methods, starting from its unified principles. The advent of computational methods by Alan Turing in 1940s and later the democratization of computational methods in 1980s created a new diversified landscape of system modeling approaches. As a result, after fifty years we encounter a competitive ecosystem of different modeling species with different capacities and trade-offs. Figure 2 shows a list of different modeling methodologies in an indexical manner.

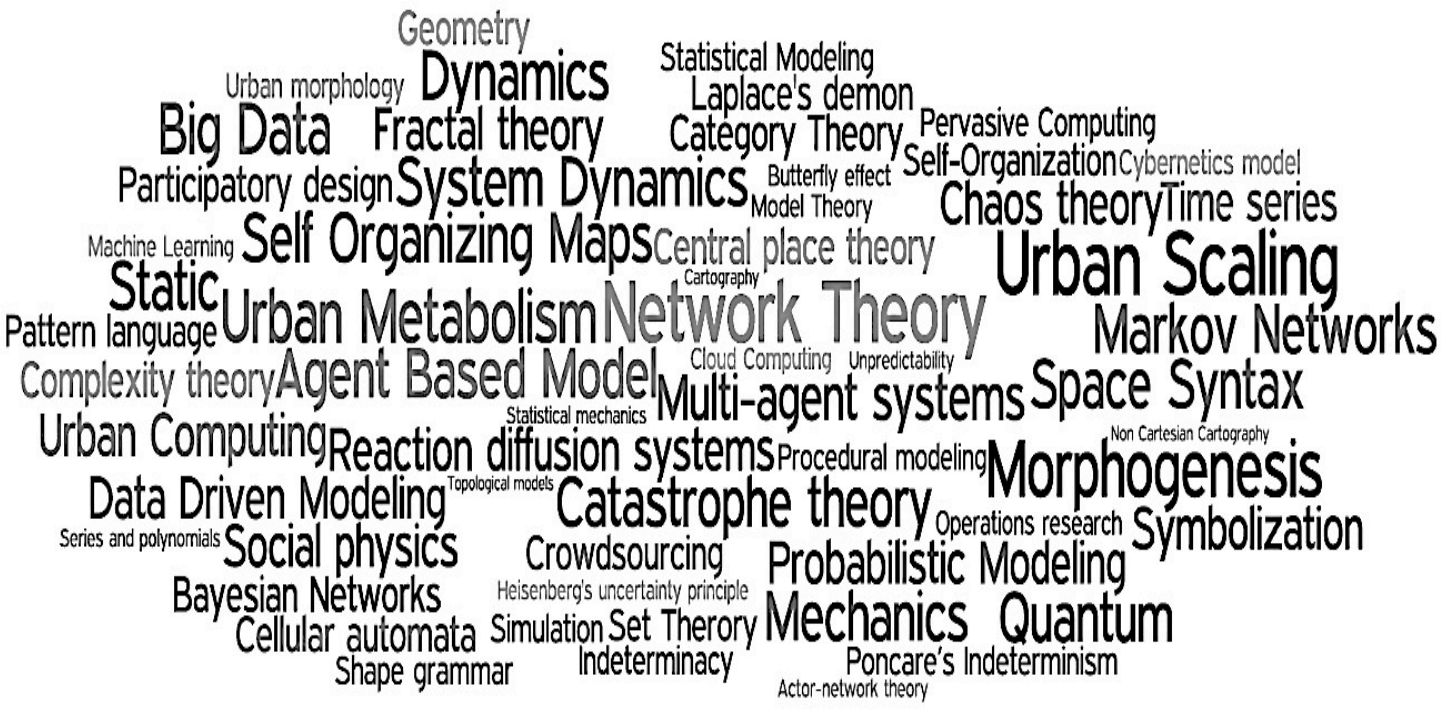

Figure 2. Competitive ecosystem of different modeling methodologies.

Therefore, the first motivation of this present essay is to find a unifying (abstract) perspective for assessment of different (inter-disciplinary) modeling approaches, while keeping the diversities. We think toward this aim we need to investigate the mathematical and philosophical grounds of scientific modeling.

## 2. Formal Definitions and Categories of Scientific Modeling

Because there is such a wide variety of modeling approaches in different scientific domains, formalizing and theorizing the practice of scientific modeling is an active research area in philosophy of science. For example, according to Roman Frigg and Stephen Hartmann, there exist the following types of models:
Probing models, phenomenological models, computational models, developmental models, explanatory models, impoverished models, testing models, idealized models, theoretical models, scale models, heuristic models, caricature models, didactic models, fantasy models, toy models, imaginary models, mathematical models, substitute models, iconic models, formal models, analogue models and instrumental models are but some of the notions that are used to categorize models.[6]
Nevertheless, these categories are not still abstract enough, but rather labels for different (not necessarily exclusive) modeling approaches.
To better understand of models, one can look at the interpretation of their roles and functions, and to distinguish the presets on which the different points of view are based. One of the main issues by which models have been extensively discussed is the relation between models and the way of representation of real phenomena under study (the target system). According to Frigg and Hartmann, from a representational point of view there are "models of phenomena" and "models of data"[7] and within these categories there are subcategories such as "scale models,"[8] "idealized models,"[9] "analogical models,"[10] like the hydraulic model of an economic system, which are further divided to material analogy, where there is a direct similarity between the properties (or relations between properties) of two phenomena, and formal analogy, where two systems are based similarly on a formalization such as having the same mathematical equations that describe both systems. (Hesse 1963). Further, one can refer to phenomenological models, which are focused on the behavior of the particular phenomena under investigation rather than on underlying causes and mechanisms.[11]
Further, there are models of theories, like the dynamic model of the pendulum, which is based on Newtonian laws of motion. Models can also be divided into ontological classes like physical objects, fictional objects, set-theoretic structures, descriptions, and equations.
However, these categories of models and modeling approaches overlap and they are rather descriptive and neutral classifications than critical. *They do not give us a measure or a gauge to compare different modeling approaches in terms of their capacities and their limits in dealing with different levels of complexity in real world problems. In this essay I am looking for a way to condition modeling approaches in different levels of complexity to examine their theoretical capacities.*
Among the above-mentioned categories, the crucial, but somewhat commonly accepted shared property of the majority of traditional scientific modeling approaches is that they are all based on some sorts of idealization.
In what follows, I explain different aspects of idealization in scientific modeling and, following the issues of idealizations, directs us to the problem of universals*,* which is an old philosophical issue.[12]

## 3. Idealization in Scientific Modeling

In the context of philosophy of science, idealization in modeling has been discussed extensively.[13] In principle, idealization is considered to be equal to an intended (over-) simplification in the representation of the target system. Although there are different ways of explaining or defining the notion of idealization, Michael Wiseberg discusses three kinds of idealization that we refer to in this work: minimalist idealization, Galilean idealization, and multiple-model idealization.[14]

Minimalist idealization is the practice of building models of real world phenomena by focusing only on main causal factors. Therefore, as is inferred from its name, minimalist models usually end to very simple elements that are informative enough for further decision-making. For example, the aim in domain of networks analytics is to explain complex behaviors that happen in the real phenomena by means of network properties such as centrality measures, integration, closeness, between-ness, etc.[15] As an example, in urban theory, in the city science approach[16] or urban scaling laws[17] the final goal is to find a few main informative factors in cities such as city size or population in order to explain other aspects of cities such as energy consumption in a linear equation. Even though it seems obvious that cities are complex phenomena with many observable aspects and many exceptions, minimalist models attract attention exactly because they identify and state very general rules.

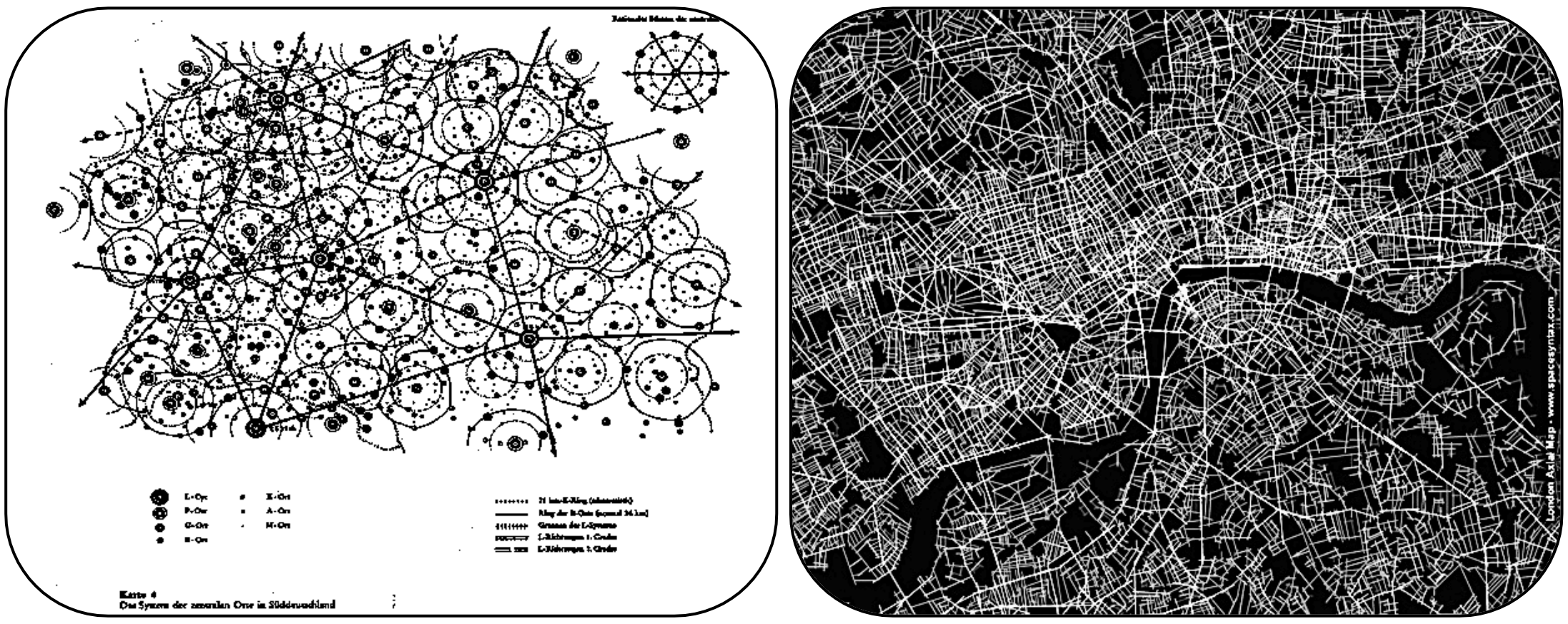

Figure 3. Network analytics: Structure oriented modeling (minimalist idealization), Central Place Theory (left) and Space Syntax (right).

City theories that seek to create archetypical city models are in a way minimalist idealized models. For example, Lynch's City of Faith, City of Machine, or City as Organism or Cedric Price's egg analogies of the city (city as boiled egg, city as fried egg, or city as scrambled egg) are characterized by few urban elements that are informative enough to explain each model and to discriminate that city model from the other models. David Grahame Shane shows how three above-mentioned models could be identified by linear combinations of three recombinant elements, called Enclave, Armature, and Heterotopia.[18]

The second category of Galilean idealization as the most pragmatic type of idealizations happens when the modeler intentionally simplifies the conditions of a complicated situation toward more computational tractability and simplicity. For example, it is common in economic models to assume that agents are rational maximizers, or in transportation models to assume that commuters take the shortest path, or to assume there is no friction in motion models of the particles. The basic idea of Galilean idealization is that by understanding the modeling environment gradually, it is possible to de-idealize or to build more comprehensive models on top of previous ones. Therefore, the majority of engineering approximation methods such as systems of differential equations or computational fluid dynamics or biological reaction networks are among this category of idealized models. Further, figure 4 shows how the idealization process in a complex phenomena (here, the agent based modeling of land-use transportation dynamics of a city) leads to a parametric and feature based representation of the real phenomena. This layering and parameterization gives the modeler the option to adjust the resolution (levels of details) of the model based on the needs and the purposes of the modeling process and the constraints and limitations, including the availability of data or prior knowledge or time and scale resolutions.

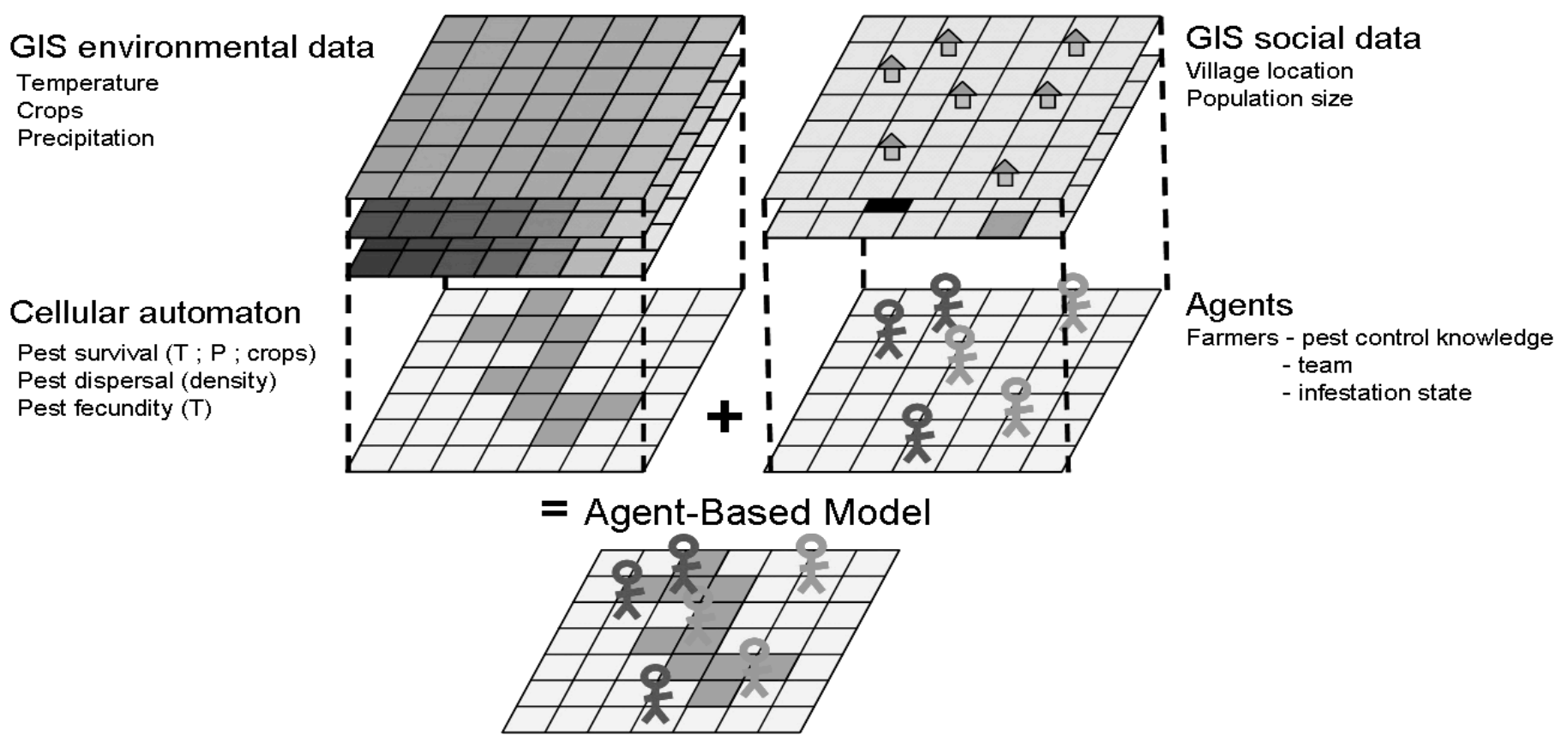

Figure 4. Parametricism: Idealization of the interactions between different agencies through layering and parameterization of the real phenomena.

The third category of idealization, multiple-model idealization, results to those models that consist of several (not necessarily compatible) models or several models with different assumptions and different properties. This type of idealization is in fact a combination of two other idealizations and it can be very useful when understanding the final output (the behavior) of the model is more important than knowing the underlying mechanisms of the target phenomena. For example, in weather forecasting, ensemble models, which (Gneiting and Raftery 2005) include several predictors with different parameters or even different structures, are used to predict weather conditions.[19]

Further, from a systemic and functional point of view there are many models in which idealization is happening in (one) main aspects of real phenomena. To just name a few: static or dynamic models, structure-oriented idealization (in network models), process-oriented idealization (such as system dynamics,[20] system of differential equations), rule-based idealization (such as cellular automata[21] or fractals[22]), and decentralized interactions (such as agent based), all are placed in the above mentioned categories of idealizations.

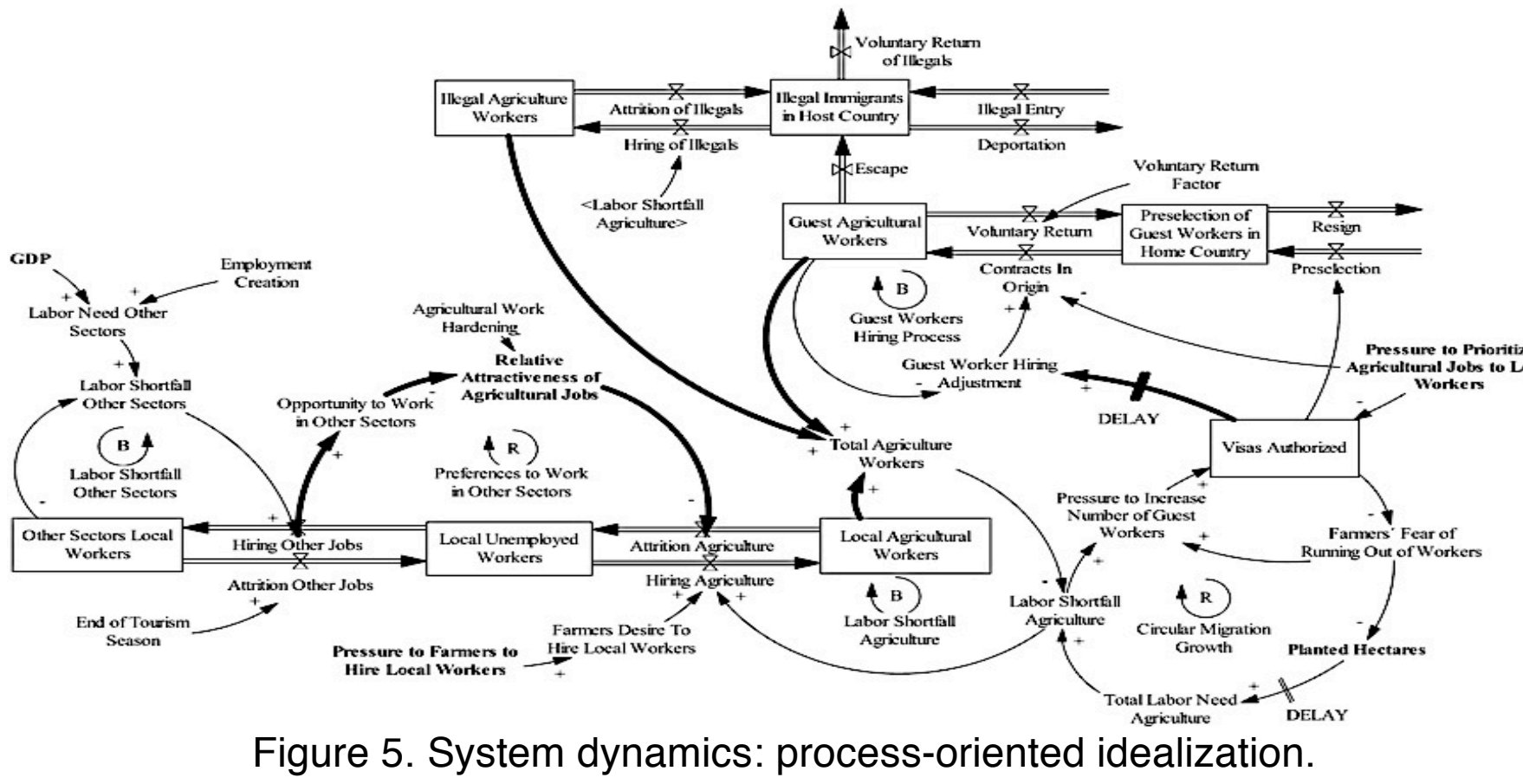

Figure 5. System dynamics: process-oriented idealization.

However, considering the size and the variety of parameters and aspects in the target phenomena, idealized models create a dichotomy, where on one extreme the models are all general, simple and tractable, and on the other, models become complicated, specific and high-resolution. In fact, multiple model idealization becomes necessary whenever the selected parameters and aspects of the target system in each individual model (out of Galilean idealization for example) are not sufficient, but also add more aspects to an individual model, either making it more complicated or resulting in model inconsistency. This issue seems to be a never-ending debate in many scientific fields including biology, ecology, economics, and cognitive and social science, where one group believe in the explanatory power of models and the other group believes in model accuracy and the level of details comparing to the real phenomena.[23]

Although idealized models have been applied successfully in many classical modeling problems, but this type of debate cannot be fruitful in dealing with complex systems as long as there is no abstraction from the current paradigm of scientific modeling (i.e. idealization). Analogically, an onion-like model of numbers explains what I mean by the abstraction in the concept of modeling. For example, with natural numbers (or more generally, integers) one can never grasp the richness of proportions and fractions in rational numbers (e.g., 2.6, which is neither 2 or 3 from a natural number perspective), while the introduction to the concept of rational numbers as the ratio of two integer numbers (e.g., 26/10) solved this problem. Therefore, by choosing 1 as the denominator, one can show that all the integers are rational numbers; while with rational numbers we have new capacities in addition to integers. Similarly, if we take an idealized model as an arbitrary representation of real phenomena by adding several of them together (which is the case in multiple model idealization), we still cannot grasp the whole complexity. Therefore, our hypothesis is that an abstraction to the concept of modeling is needed in order to conceptually encapsulate all the potential arbitrary views in an implicit way.

However, I do not claim that one can introduce a new concept as such, but in fact in this work I am trying to identify and discover new aspects of a potential body of thinking in scientific modeling.

In order to highlight this conceptual abstraction from the current idealization paradigm, first we need to explain the notion of universals, including abstract and concrete universals, followed by our interpretations of these concepts in relation to the notion of scientific modeling.

In the next section, after presenting the connections between the notions of idealization and abstract universals, I will formally describe the concepts of abstract universals and concrete universals, which can be interpreted as set theoretical and category theoretical definitions of these two notions.[24] Further, I will show how the concept of concrete universals from category theory can open up a new level of modeling paradigm.

# 4. Universals and Modeling

In the majority of texts written about idealization in the domain of scientific modeling, the notion of idealization is equal to simplification and the elimination of empirical details and deviations from a general theory that is the base for the final model. At the same time, the word “ideal” literally comes along with “those perfections that cannot be fully realized.” For example, circle-ness as a property is an ideal that cannot be fully realized, and any empirical circular shape has, to a degree, the circle-ness property.

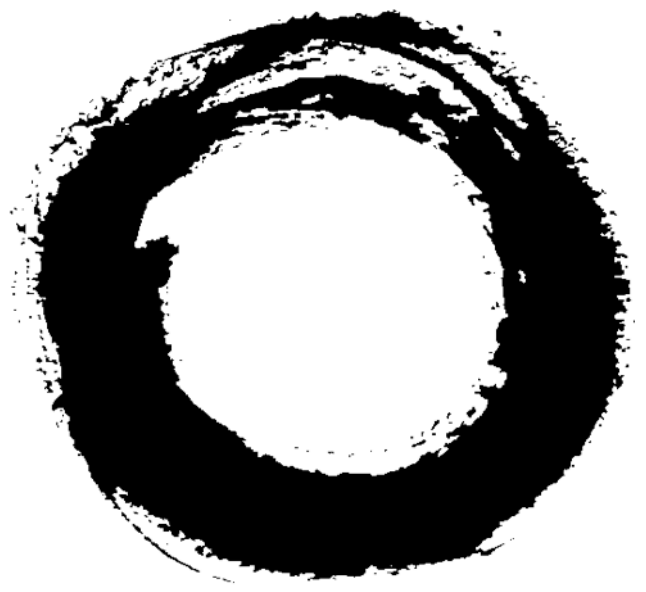

Figure 6. Enso (circle of Zen): Toward the ideal circle.

Therefore, the idealization process in scientific modeling can be explained as a form of purification of empirical observations toward a set of given (assumed) ideal properties. In statistical data analysis, it is always assumed that collected empirical data follows a normal distribution function. Thus, one can convert the empirical data to a normal distribution function and utilizes from the machinery of this ideal mathematical representation (i.e. the normal distribution function). Applications of idealizations in many mathematical approaches such as linear algebra are enormous. For example, a Fourier transformation (Figure 7) can be seen as a form of idealization by which any observed time-varying data can be reconstructed (approximately) by a set of time-varying vectors (a set of pure sinusoidal waves with different frequencies and phases). From this perspective, any waveform phenomenon is a linear combination of a set of ideal prototypes.

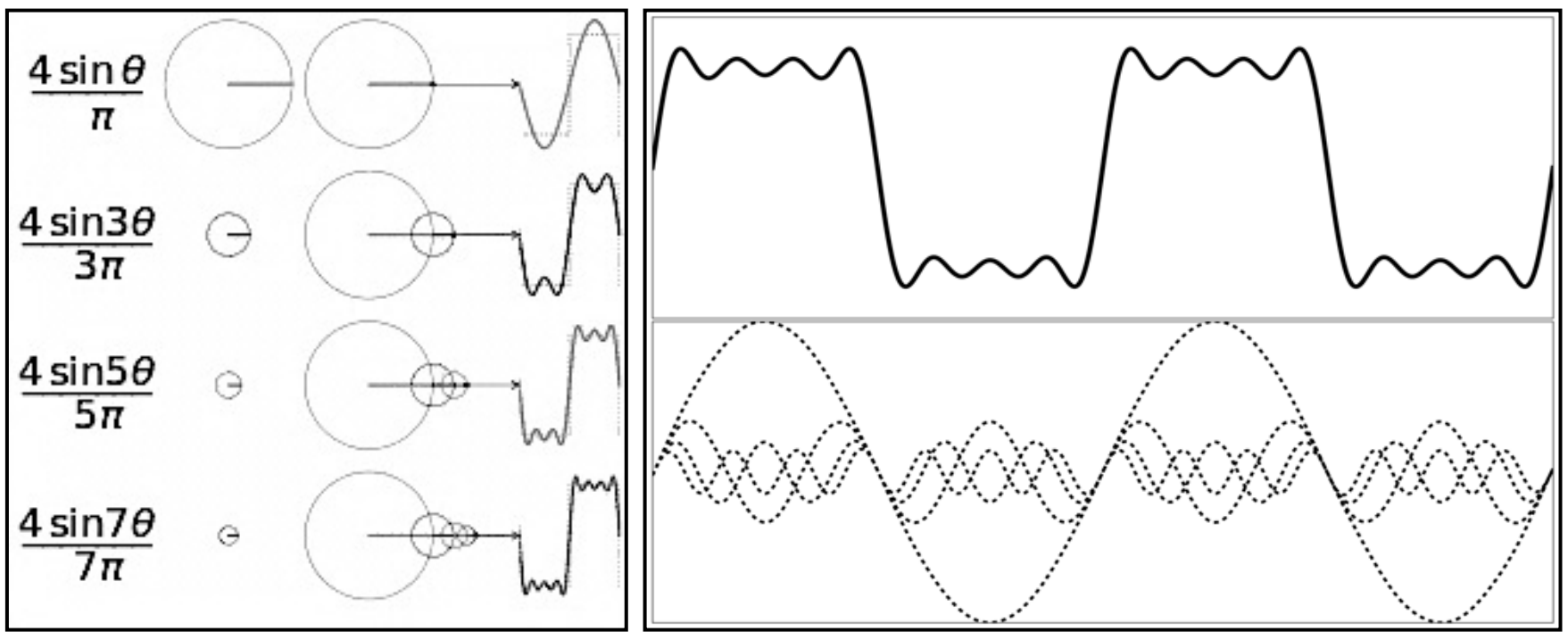

Figure 7. Fourier decomposition: Any observed form is a linear combination of some ideal cyclic form.

However, these ideal forms (a wave with a certain frequency in the case of the Fourier analysis) as the set of aspects (properties) of real phenomena are abstract. This means that there is no concrete (empirical) instance that fully matches one or several of these a priori, ideal properties. From this point of view, idealized models are models that are based on the notion of abstract universals.

The notions of “universals” and “property” are old topics in philosophy that can be approached differently, namely through realism, idealism, or nominalism.[25] However, in this work I focus on the distinctions between concrete and abstract universals in relation to the paradigms of scientific modeling.

According to David Ellerman, “In Plato's Theory of Ideas or Forms (ειδη), a property F has an entity associated with it, the universal $u_F$, which uniquely represents the property. Therefore, an object X has the property F i.e. F(X), if and only if it participates in the universal $u_F$ to a degree ($\mu$).”[26] For example, “whiteness” is a universal and the set of white objects that participate in “whiteness” property (i.e., with different degrees of whiteness) are represented by this property. Further, “Given a relation $\mu$, an entity $u_F$ is said to be a universal for the property F (with respect to $\mu$) if it satisfies the following universality condition: For any x, x μ $u_F$ if and only if F(x).”[27]
This condition is called universality, and it means that the universal is the essence of that property.
In addition to universality, a universal should be unique. “Hence there should be an equivalence relation (≈) so that universals satisfy a uniqueness condition: If $u_F$ and $u_{F'}$ are universals for the same F, then $u_F \approx u_{F'}$.”[28] Therefore, any entity that satisfies the conditions of universality and uniqueness for a certain property is a universal for that property. Now, if a universal is self-participating, it is called a concrete universal; if it does not have self-participatory properties, it is an abstract universal. For example, whiteness is an abstract universal as there is no empirical (concrete instance) to be “whiteness.” In language models, being a “verb” is a property that can be assigned to many words, but “verb” itself is an external definition and it is not self-participating in the sets of concrete verbs. The same argument goes for the above example of the Fourier analysis and ideal forms.
On the other hand, defining a property as being part of set A and set B has a concrete universal, which is the intersection of two sets A and B (A∩B). It means that any object from set A and B (including all the potential subsets) that has this property (being part of A and B) participates in the intersection set A∩B, and since A∩B is participating in itself, then it is a concrete universal.
Further, Ellerman shows that how modern set theory is the language of abstract universals and how category theory can be developed as the mathematical machinery of concrete universals. Finally, he summarizes that, Category theory as the theory of concrete universals has a different flavor from set theory, the theory of abstract universals. Given the collection of all the elements with a property, set theory can postulate a more abstract entity, the set of those elements, to be the universal. But category theory cannot postulate its universals because those universals are concrete. Category theory must find its universals, if at all, among the entities with the property.[29]
In the past few decades there have been many theoretical works to further the new field of category theory in terms of this fundamental difference between set theory and category theory. For example, currently the main categorical approaches in mathematics are topos theory and sheaf theory, which are generalizations of topology and geometry to an algebraic level.[30] It seems that applications of these general frameworks in different domains should be one of the main future research areas in the field of modeling. On the other hand, Ellerman concludes that, Topos theory is important in its own right as a generalization of set theory, but it does not exclusively capture category theory’s foundational relevance. Concrete universals do not “generalize” abstract universals, so as the theory of concrete universals, category theory does not try to generalize set theory, the theory of abstract universals. Category theory presents the theory of the other type of universals, the self-participating or concrete universals.[31]
Now that we have defined the concepts of abstract and concrete universals, we need to formalize two different approaches of modeling, which are based on these notions of the universal.
As stated earlier, idealized models are models that are based on the notion of abstract universals and consequently idealized models can be interpreted as set theoretical models. In the next section, by focusing on the idea of representation in idealized models, I show their theoretical consequences and their limits in dealing with complex systems, with the definition of the abstract universal being crucial. Next, I show another conceptual representational framework that is matched with the concept of concrete universals. Further, I will introduce an alternative line of modeling to idealized modeling.

# 5. Specific Modeling: Models Based on Abstract Universals

The fundamental difference between abstract and concrete universals is the issue of self-participation. In terms of modeling and representation, in those models based on abstract universals, the definition of the common property of the target system is a priori, given in a meta-level. This means that in an empirical setup, we have an externally given idea about the set of properties (aspects) of the real phenomena under study at the beginning of the modeling process. As an example, if we are comparing many concrete objects (e.g., several apples), we first need to define a set of specific properties (such as size, color, taste, etc.) to construct a representation of apple-ness. Therefore, apple-ness is reduced to this external setup. We call this approach specific modeling as it is based on a set of specific properties of the target system. In relation to the idealization process, the level of details in terms of the number and variety of properties is the choice of the modeler. If the modeler considers few aspects of the target system the model becomes simple and if he or she selects many aspects or properties, the model becomes complicated.

Figure 8 shows the concept of idealized representation in specific modeling schematically. Each circle in this figure stands for a concrete object. These objects are symbolical, which means that they can stand for anything—be it people, cars, companies, buildings, streets, neighborhoods, cities, webpages, protein networks, networks of words in a corpus of texts, or people and their activities in a social network. Therefore, in the first step, we need to define our abstract universals, which leads to a set of selected features of the real objects. These features are shown by rectangles. As a result of these universal features, the concrete instances of the object are assumed to be independent from each other, as they will be all compared indirectly by an abstract class definition, which acts as an external reference.

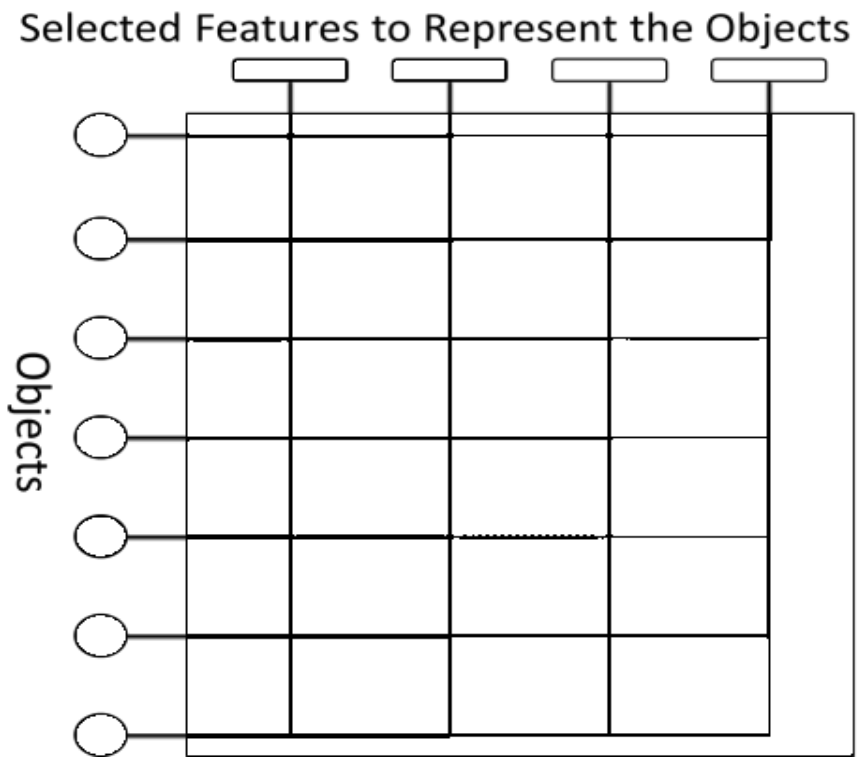

Figure 8. Specific modeling-based abstract universals and parametric idealization of the target object.

This is the underlying notion of rationality started in sixteenth century by René Descartes and it should be mentioned that it offers a fantastic mechanism and an abstract language for axiomatization of different phenomena. Nevertheless, there are fundamental limits to this approach of modeling in dealing with complex phenomena, with many different properties, where the specific models need to define an arbitrary set of properties.

## 5.1. Limits of Modeling Based on Abstract Universals

Within the literature of scientific modeling, the majority of discussions on the issues of scientific modeling approaches are bounded to models based on abstract universals and the differences of different idealization processes. Among the few investigations, Richard Shillcock discusses the fundamental problems of modeling in the domain of cognitive science from the perspective of universals. He notes: "Cognitive science depends on abstractions made from the complex reality of human behaviour. Cognitive scientists typically wish the abstractions in their theories to be universals, but seldom attend to the ontology of universals."[32] Later he explains several fundamental problems in the domain of cognitive science by reviewing the different aspects of abstract and concrete universals. In what follows I present some of the fundamental issues of the models that are based on abstract universals.

#### 5.1.1. Godel's Incompleteness Theorem and Arbitrariness of Models Based on Abstract Universals

In models based on abstract universals, the universal properties are not self-participating. Intuitively, one can argue that in any level of abstraction, members of a set are concrete and the set itself is abstract with regard to its members. Therefore, the first modeling step is the decision about the set of properties that define (represent) the object of enquiry. To have a set of concrete instances (e.g., set of red apples), one needs a super-set that defines the ideal properties of that class (the apple-ness and the red-ness). This requirement (brought forward by Plato) initiates a never-ending hierarchical process of defining abstract universals for the higher order classes (e.g., a set for colors). As a result, one can argue that in practical modeling domains, from a level above, models are based on assumed or commonly agreed properties of the target system. This problem can be explained by Gödel's incompleteness theorem, that is to say we only can make a consistent system if it is based on an unproved truth (the incomplete model); if the model is complete (every thing based on proofs), it cannot be consistent.[33] This beautiful theorem simply says that any model that is based on abstract universals is in a way arbitrarily consistent, but not simultaneously complete. The same argument holds for the case of Russell's paradoxes and naïve set theory.[34]

#### 5.1.2. Curse of Dimensionality in Complex Systems

Models based on abstract universals have been successfully applied in many practical domains such as classical physics, medicine, and engineering. Nevertheless, they reach a computational limit in dealing with complex systems. This limit is directly related to their quest for explicit representation of the target systems through a set of specific properties. Assume that we measure the complexity of a system (i.e., a real phenomenon) as a function of the number of its potential properties and the relations between those properties.[35] In this scenario, in comparison to a building a wooden chair is less complex, and the same relation holds for a building in comparison to a city.

As a result, by increasing the number of potential properties and their interrelationships, and consequently the exponential growth in the number of combinations, the space of modeling (i.e., potential specific models) expands in an exponential manner. This phenomenon is called the curse of dimensionality introduced by Richard Bellman in 1961.[36]

Consequently, in a complex system, any endeavor toward an explicit representation (which is the case in specific modeling) leads either to a complicated model (models with lots of redundancy and lack of explanation) or to very simple and minimalistic idealizations. Figure 9 shows this issue diagrammatically.

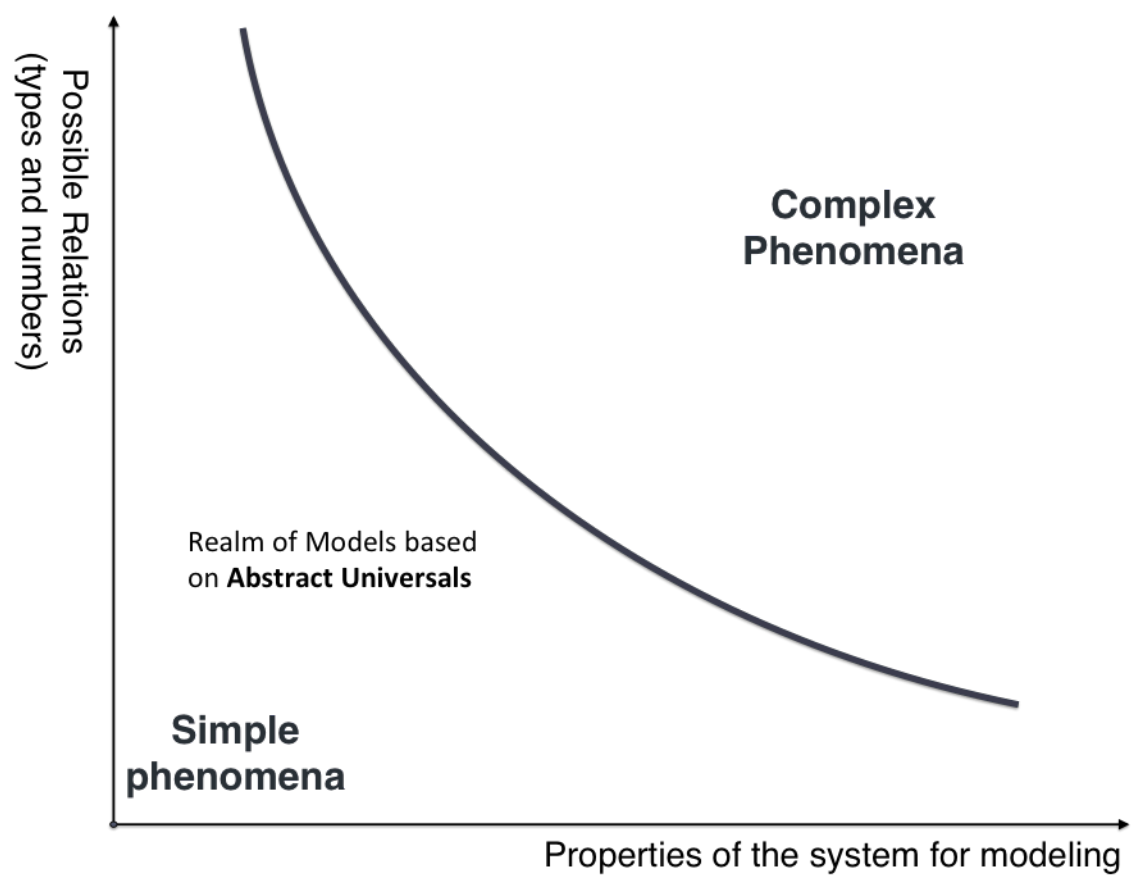

Figure 9. The curse of dimensionality and idealized models based on abstract universals.

### 5.1.3. From Particular to Generic and the Concept of "Error"

In the idealization process of particular objects there is no longer a unique identity dedicated to a particular (concrete) instance, but rather the identity of that particular case is realized as a combination of globally defined properties (see figure 8). In other words, in models based on abstract universals the particular object is considered as an instance of a (fictitious) generic object. Along the same line, Shillcock says: "The concrete universal is a universal, but it has all the richness of the particular. Whereas an abstract universal can be defined as something abstract (typically seen as a property) that inheres in many other different things, a concrete universal is an entity in which many other different things inhere."[37]

Consequently, constructing the notion of generic object through the lens of abstract universalism, we impose a limit to empirical deviations and treat them as errors. For example, assuming linearity as an ideal property in a system, the generic object is assumed to follow the linearity while other objects are erroneous to a degree, based on their deviations from the linear line. Figure 10 shows this issue in the case of linear regression. In a two-dimensional linear system, it is assumed that for any observation (i.e., a concrete object via its x and y dimensions) there is a linear relation in the form of $y=ax + c$. Therefore, those points that don't fall in a common line have a degree of error in comparison with the ideal line.

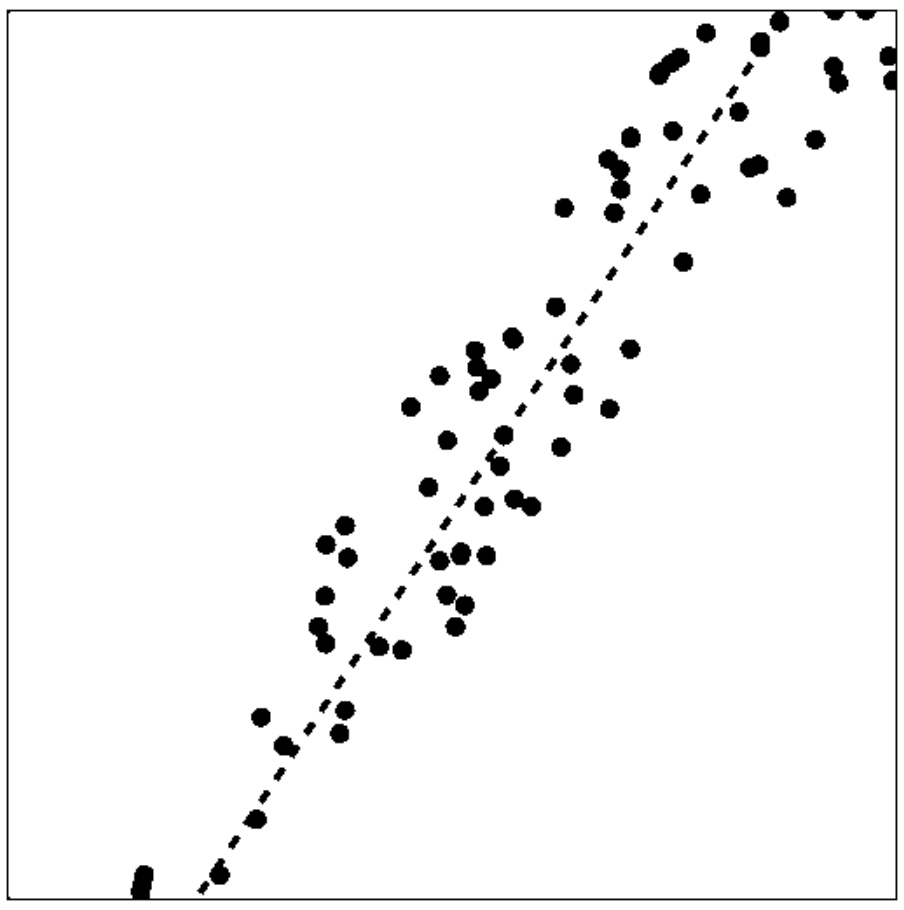

Figure 10. Introduction to the concept of error: the deviation of particular objects from the ideal line

# 6. Pre-Specific Modeling: Models based on Concrete Universals

In this section, we investigate the potentials for a new level of abstractions in paradigms of scientific modeling. This is not in opposition to specific modeling (i.e., models based on abstract universals), which is a common approach in social science and humanities, rather it is based on the notion of concrete universals from category theory. As discussed in the previous section, models that we call specific models are based on a priori defined or selected abstract universals. Further, they have certain theoretical limits and issues in dealing with complex systems. Therefore, my hypothesis is that if any specific model is like an arbitrary view to the real phenomena, there should be a category of models that encapsulates all the potential specifics in an implicit way. We call this approach pre-specific modeling, which was originally introduced by Vera Bühlmann.[38]

If specific modeling can be theorized by set theory and abstract universals, pre-specific modeling should be supported by the concepts of category theory and concrete universals. In order to establish the building blocks of pre-specific modeling, we need to focus on fundamental assumptions of the specific modeling.

## 6.1. Dedekind Cut: When a Particular Object is Represented by the Negation of Its Complement

In specific modeling, when one defines the abstract universal in terms of a set of specific properties, a parametrical generic object will be conceptualized directly. The individual objects can then be reconstructed (analyzed) or generated (synthesized) by changing the values of those specific parameters in the generic object. As a fundamental example, we refer to number theory and the definition of rational numbers as the ratio of two integers m and n, where n is not equal to 0. In this case, any specific rational number, q, can be directly represented by infinite pairs of (m, n) integer values, where m/n = q. In other words, q is graspable directly and independently from the other rational numbers. However, as we know the set of rational numbers are countable and they are only a small fraction of the whole space of the real numbers. As a result, this approach reaches to a point in some cases that no one can define a real number by its own.

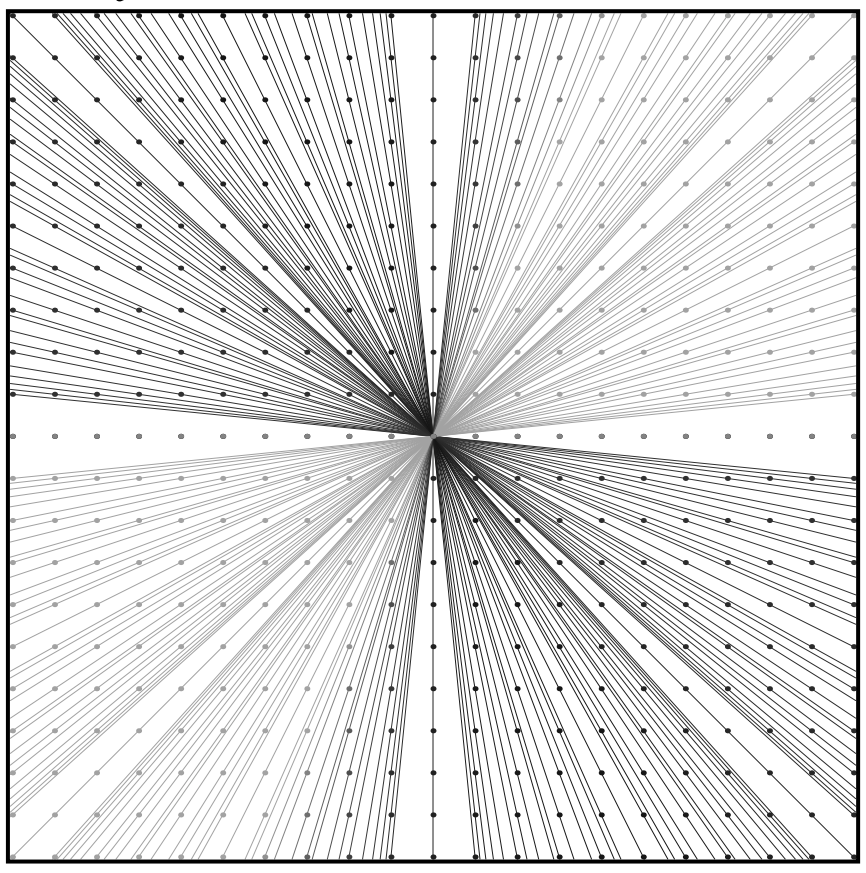

Figure 11. Rational numbers cannot fill the space of real numbers. Each line corresponds to one rational number.

For example, √2 cannot be touched by the above-mentioned procedure. In general, this is the case for irrational numbers. A different method for definition of irrational numbers is required. In the late nineteenth century, Richard Dedekind came up with a different conceptual definition of irrational numbers, known as the Dedekind cut.[39] Intuitively, a Dedekind cut is a unique way of representing an irrational numbers by its complementary set. He defined a cut for a specific number, b, as the space between two ordered sets of rational numbers A and B, where all the elements of A are less than all the elements of B and further all the elements of A are smaller than b and all the elements of B are equal or greater than b. Definitely, if b is a rational number the union of set A and B is the whole number space of real numbers, U, and if b is an irrational number, b is equal to U minus the union of sets A and B (AUB). For example, in order to define √2, A is the collection of all negative rational numbers and the collection of every non-negative rational number whose square is less than 2 and B is the collection of all positive rational numbers whose square are larger than 2.

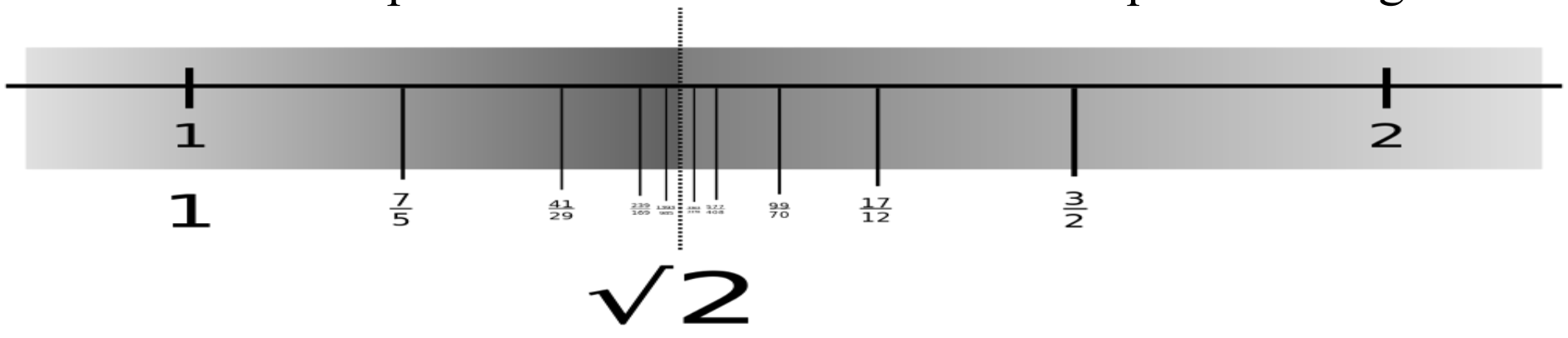

Figure 12. Dedekind cut: representation of an irrational number as the negation of its complement

Further, two irrational numbers can be compared through their corresponding cuts—if their cuts are equal, these numbers are identical. By this definition each specific irrational number is represented uniquely as the negation of its complement, while it is not directly touchable. This is opposite to that of a rational number, where it can be directly pointed out. We think that regarding the issue of object representation, the Dedekind cut importantly implies that it is possible to introduce an alternative approach for representation of the objects to what is common in specific modeling (shown in figure 8).[40]

**6.2. From Generic to Particular: Object Dependent Representation**

The core aspect about any modeling paradigm is how real phenomena are represented. As figure 8 shows, by selecting the set of representational properties of the real phenomena in specific modeling, each individual object is represented directly. In other words, the identity of a particular object is defined independently of the other (concrete) objects as long as we have a global axiomatic set up (i.e., those selected properties) to define the generic object. Here, the generic object is the abstract universal, in which with different parametric values one can instantiate or approximate a particular object. Referring to the example of number theory, this is similar to the case of rational numbers, where a specific number can be generated as the ratio of two integers.

Now, imagine an empirical representation of concrete objects in network-based representation in which nodes of connectivity, which are to specify in multidimensional ways, represent objects. For example, the number of cars that pass from one street to another, or the relation established by two individuals who select the same restaurant, or the relation between two cities that host offices of the same company, or the number of times a specific word has appeared after another specific word. In distinction to parametric representation of objects, the identity of an object is defined directly in terms of the relations it maintains with the other objects. The main difference between the two approaches is that in the feature (property) based approach, the specific identity of objects is assumed independently, while in the network-based representation, the identity of objects is regarded as pre-specific, and is specified purely relationally, out of the connectivity, which is observable. Two objects are considered identical if they share the same sets of relations with other objects.

Figure 13 shows the representation of concrete objects in a network-based approach.

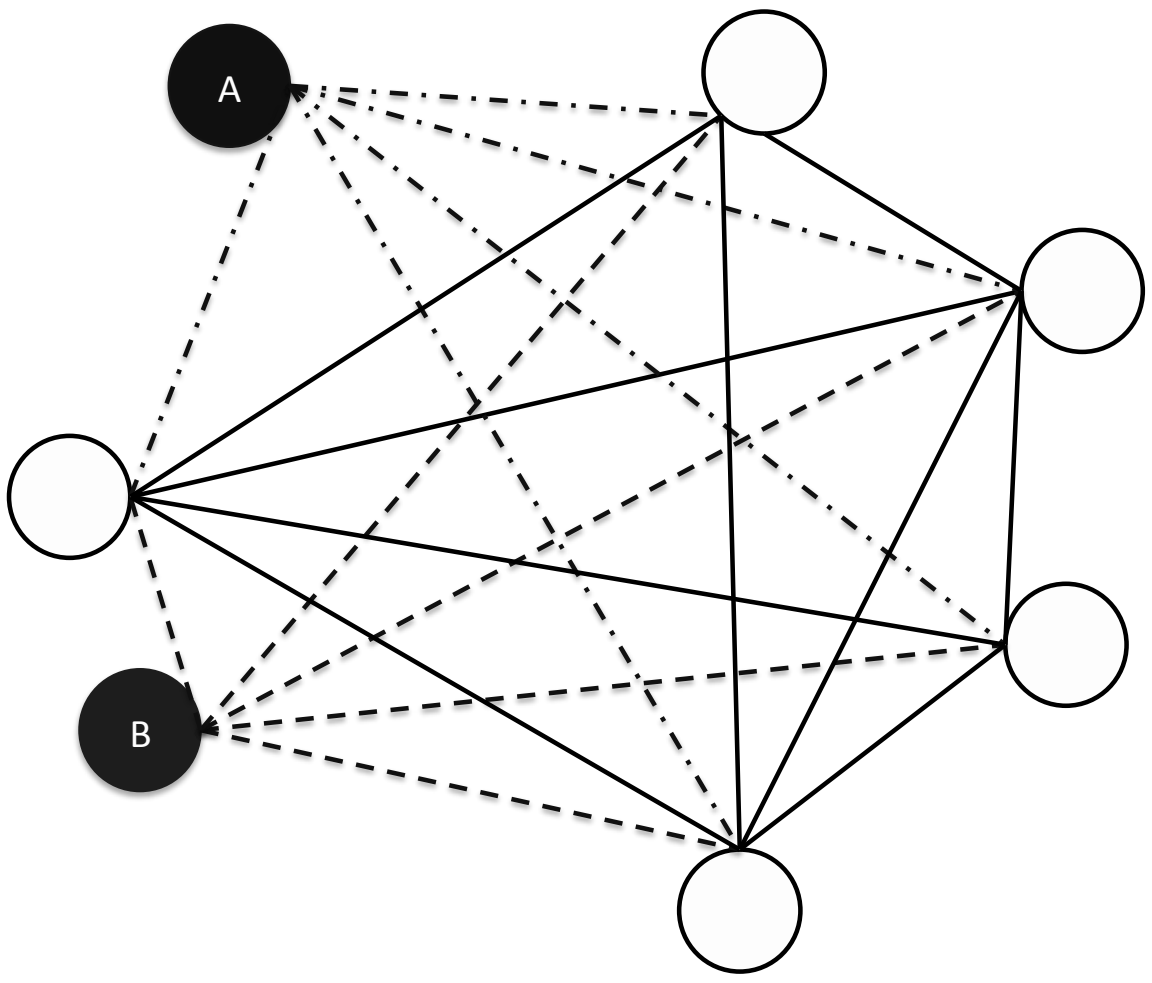

Figure 13. Object-dependent representation of concrete objects in pre-specific modeling

In specific modeling, each property has an abstract universal, but in object-object relations, each concrete object is a property (for example A-ness for concrete object A) and thus we have concrete universals for each property, as each object has an identity relation with itself. Assuming each concrete object as a feature, we can represent each object via its relation to the other objects. In comparison with the definition of irrational numbers by the Dedekind cut, here too the identity of a particular concrete object is defined as the negation of the identity of the other particulars. Note that here there are not yet defined generic objects—unlike the case of parametric object representation.

This set up as shown in figure 13, is an object-dependent representation that is conceptually scalable with the size of empirical objects. While in specific modeling the size of parameters is independent to the number of concrete objects (i.e., observations), in object dependent representation, by adding one concrete object we directly add one new aspect for the representation of other objects. This aspect makes pre-specific modeling suitable for working with large amount of data. Further, in many areas today, the conditions for these types of representation hold, as we have an emergent network of connected instances that can be used for the representation of the object of inquiry.

In section 8 I present two main technical frameworks that support the concept of pre-specific modeling—two main applications of object-dependent representation. However, as I mentioned before implicitly, object-dependent representation and pre-specific modeling in general are data driven, as the setup shown in figure 13 is based on concrete objects. The role of data in pre-specific modeling is different than classical empirical research when one assumes to have an a priori generic object. As pre-specific modeling is proposing a new modeling framework it demands another notion of data, one that is different than traditionally designed observations and measurements.

## 7. Massive Unstructured Data Streams: An Inversion in the Notion of Measurements and Data Processing

In classical scientific modeling, theories and a priori representations define what should be measured and observed. According to Bas C. van Fraassen, A measurement outcome is always achieved relative to particular experimental setup designed by the user and characterized by his theory.[41] Similarly, as we showed in the case of specific modeling, by selecting abstract universals we limit the set of potential observable aspects of real phenomena. For example, when dealing with a pendulum model and using Newtonian laws of gravity as a theoretical model to describe the foundation of the motions of particles, data and measurements can only empirically validate or propose minor modifications. Therefore, classical data has always played a marginal role in the process of modeling.

In addition to this conceptual setup, measurement and observation have, historically, been very expensive, that have pushed modelers toward more structured and designed and optimized experiments and observations. Taking into consideration data in specific modeling, figure 14 shows the classical process of modeling. As shown in the diagram, abstract universals (or the definition of the generic objects) are always the first and the primary element of modeling processes; the data—including its structure (i.e., the selected properties of the real environment) and its size (to be statistically enough)—has a supporting role in model tuning and model validation. This diagram shows that since the data is the secondary element, after a certain level of observation the model quality (in terms of accuracy for example) becomes stable, as we have enough data to tune the system.

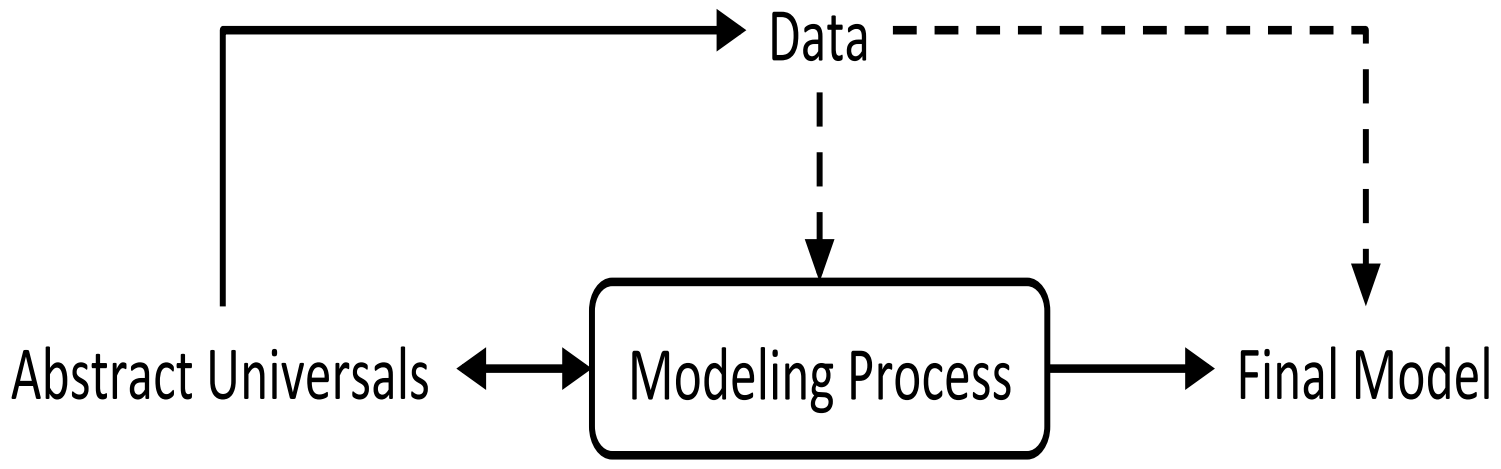

Figure 14. The classical modeling process (specific modeling)

Nevertheless, considering computational technologies as the dominant factors in shaping and directing the area of scientific modeling of the last century, the landscape of measurements and data processing has been changing dramatically. In (Moosavi 2015)[42], I discuss three levels of computational capacities, known as computing power, computational and communicational networks, and data streams. The first level deals with computing power in terms of numerical simulations in comparison with analytical approaches. Historically, there have been different technologies of computation starting with mainframes, moving to the democratization of computing through personal computers and microcomputers, which are still getting faster and more powerful at an exponential rate. The primary function of computing power is numerical simulation, even though computers have been isolated or with limited communication abilities. Although computers and their simulation power opened up new possibilities for better understanding of the real world phenomena in 1960s and 1970s in many fields, for a while during late 1970s, these computational models got data hungry and their demand for data was higher than what was available for model tuning and validation. This produced some skepticism about the applications of computational models to real world problems.[43]

However, alongside the developments within computing technologies, advancements in communication technologies gradually opened up another capacity for modelers, which can be considered as the second level of computational capacities. In this level while computing power was given, what was important was the communication between computing systems. Therefore, new phenomena such as networks of sensors, mobile phones or computers, and the Internet started to emerge. Gradually, considering the amount of embedded systems in many real world applications, computers as computing machines became the ground to introduce new functions that were emerging on top of computational networks.

As a byproduct of these networks of computing and communicating machines, the amount of digital data started to increase as well. Starting from mid 1990s, technical terms such as “data mining” and “database management” emerged in parallel to a focus on methodology to explore digital data (mainly structured data) among modelers. As can be seen, data began to emerge around this period, but this data was a byproduct of designed measurements and sensory systems. It is important to note that, by this time the notion of data had not changed from its old notion—collected data was still structured and followed by the modeler’s choices. In fact, the data still was the secondary element, rationally determined by the given properties of the target system. However, what had changed dramatically was the amount of digitally collected data. It started to grow quantitatively on top of the communicating and computational networks across disciplines.
Finally the third level, for which we think we have a suitable notion of data for pre-specific modeling, emerged only recently. With rapid advancements both on the level of computing power and the networks of computing systems, and a rapid growth in social media, we have encountered a new stage in which on top of ubiquitous computing and communicating systems, a new level of abstract phenomenon has started to emerge. We have begun to experience exponential growth in the amount of information available, together with the mobile computing devices most people use on a daily basis. This is often called a data deluge. Next to the challenges these changes bring, we can also see how new areas for research and practice are emerging.[44,45]
It seems clear today that the classic paradigm of observation and data gathering has changed radically. Data is produced on an everyday basis, from nearly any activity we engage in, and accumulates from innumerable sources and formats such as text, image, GPS tracks, mobile phone traces, and many other social activities, into huge streams of information in digital code. These unstructured and continuous flows, which can be called urban data streams, can be considered as a new infrastructure within human societies.
This notion of data is opposed to its classical notion, where data was produced mainly as the result of designed experiments to support specific hypothetical models or when data was transmitted via defined semantic protocols between several inter-operating software. These new data streams are the raw materials for further investigations; and similar to computing power, they hold new capacities for modeling. As a result of this new plateau, we are challenged to learn new ways to grasp this new richness.

These massive, unstructured urban data streams induce an inversion in the paradigm of modeling from specific modeling, and they match the concepts of pre-specific modeling and models based on the concept of concrete universals.
Therefore, as an alternative to previous modeling process, pre-specific modeling is mainly based on the coexistence of unstructured data streams representing particular objects and self-referential representation of concrete universals (figure 15). As shown in the next section, opposite to a specific modeling paradigm, where data has a limited use in modeling process and after a certain level of data size, the performance of models become stable in this level of data-driven modeling and object dependent representation. Adding more data will improve the quality of the final model. This is the power of concrete universals that are scalable with data size, unlike parametric models, in which after a certain size of data, the parametric state space of the generic object becomes full and reduces the discrimination power of the model in dealing with variety of circumstances in complex systems. By using concrete universals, each new instance will introduce a new aspect in relation to the other concrete instances. Therefore, it is scalable with data size.

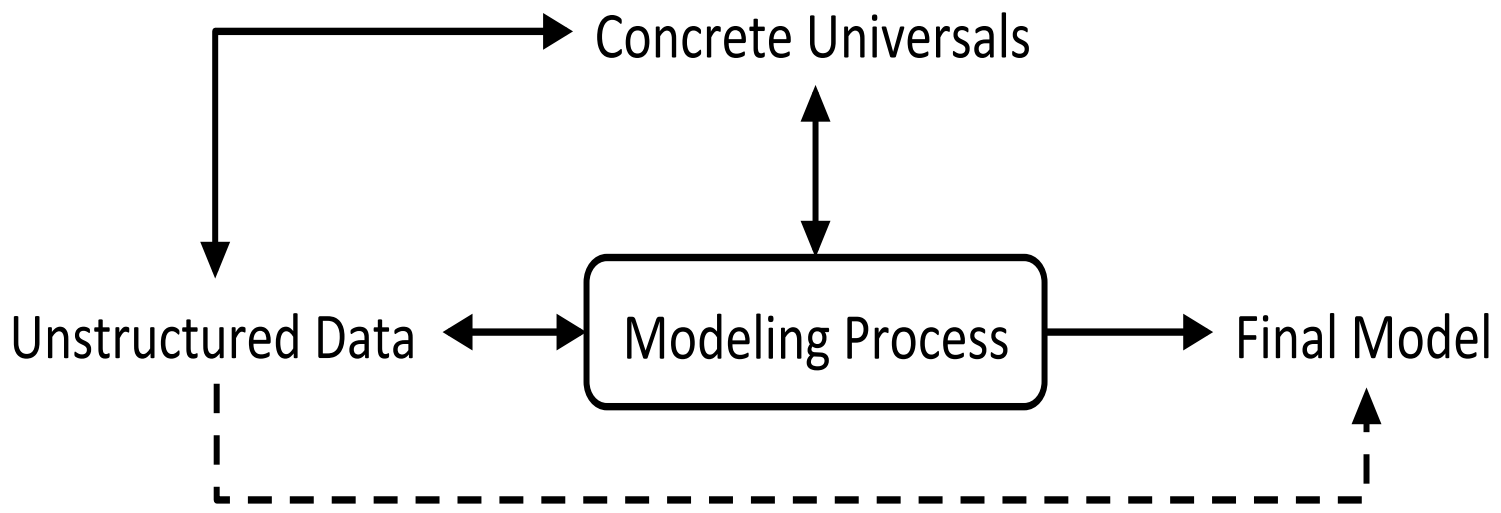

Figure 15. Pre-specific modeling in coexistence with unstructured data and concrete universals

In the next section, I present a certain category of mathematical and computational methods, which support the notion of pre-specific modeling.

# 8. Computational Methods Supporting Pre-Specific Modeling

Even though I've presented several examples of pre-specific modeling, so far the explanation has remained on a conceptual level. Here I present a more technical discussion of two computational methods that fit very well with the concept of pre-specific modeling.

**8.1. Markov Chains**

Andrei Markov is among the greatest mathematicians of twentieth century; he has made numerous contributions toward forming probability theory, but his major work is the concept of Markov chains, which he introduced in 1906. In engineering and applied scientific domains today, many people know Markov chains as a kind of memory less dynamic model, where having a sequence of random variables ($x_1$, $x_2$, $x_3$,…, $x_{t-1}$, $x_t$) the state of the system at the next step ($x_{t+1}$) depends only on the previously observed state ($x_t$). These processes create a chain of random activities, where there is a probabilistic link between adjacent nodes. Here, we assume the case of discrete time processes with a finite number of states, but in principle one can assume to have continuous time and continuous state space. Further, the chain is called homogeneous if the conditional distributions of $x_{t+1}$ given $x_t$ were independent of time steps. And, assuming more sequential dependency, one can construct higher order chains from the state at step t depends on its n previous steps, if the order of the chain is n. Nevertheless, in the case of first order chains, assuming the mutual dependencies between any two potential states, a Markov chain can be represented with a directed graphical network, where each node corresponds to a state and the edges between two nodes correspond to two conditional probabilities. Figure 16 shows an example of a Markov chain with three states.

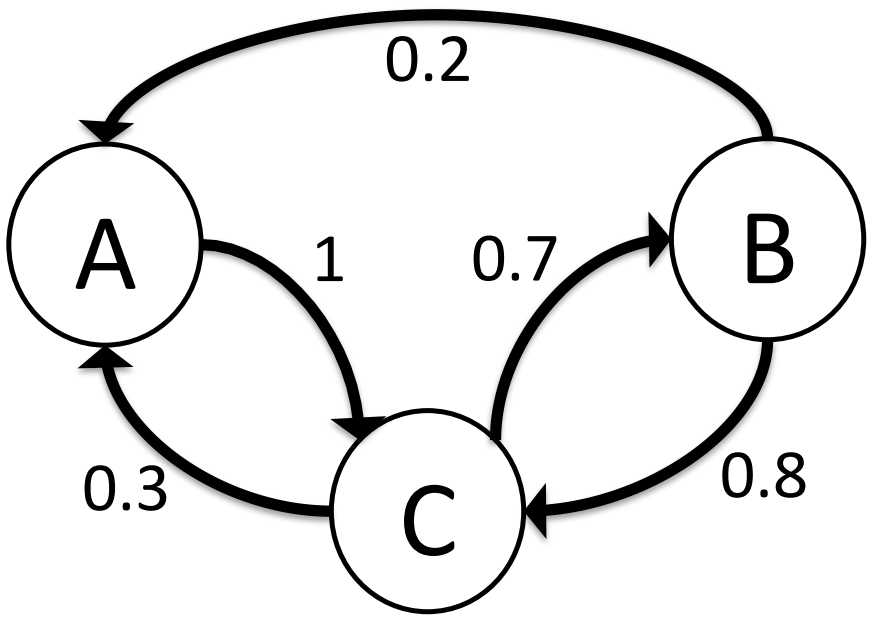

Figure 16. Traditional representation of a Markov chain in modeling of a dynamical system

In the domain of dynamical systems, Markov chains and their many different versions have been studied and applied in diverse applications for the simulation of dynamical systems or for the study of steady state conditions. They have also been applied successfully to sequence and time series prediction and classification.
Markov's brilliant idea for the representation of a complex phenomenon such as natural language in a purely computational manner is particularly relevant here. Before going into his approach, let us unpack the traditional concept of language models. One of the major and to some degree dominant concepts of linguistic models is based on the notion of abstract universals. In this approach to modeling, which is in accord with Noam Chomsky's,[46] a spoken language can be modeled by means of a set of semantic and syntactical laws of the that specific language. Therefore, writing and speaking correctly by an individual means that there is a system of production in his or her mind that produces the instances of that language following his or her ideal model. This is one of the best examples of specific modeling based on the concept of abstract universals. However, Shillcock notes that because natural languages are complex evolving systems, trying to identify the ideal model of a live language is always a process of catch up.[47] Therefore, considering the evolution and the exceptions and the number of different languages all over the world, this approach has never been successfully applied in a computational model.
Now let us refer to the experiments of Markov in 1913, which is among the first linguistic models that follow the concept of concrete universals.
In what has now become the famous first application of Markov chains, Markov, studied the sequence of 20,000 letters in Alexander Pushkin's poem "Eugeny Onegin" to discover the conditional probabilities of sequences of the letters in an empirical way. What follows is the less discussed way of interpreting Markov chains, which is not from the traditional viewpoint of dynamical systems, but is rather about the empirical representation of concrete objects. Figure 17 shows the underlying concept of object dependent representation in Markov chains. Suppose that we have a defined number of symbols in a specific language (e.g., all the observed words in the English language). Now imagine for each specific word in the collection of our words, we consider all the words that have appeared N step before and N step after that specific word in all of our collected texts and we count the total number of occurrences. Next, by normalizing the total number of occurrences for each position before and after each specific word, we can find the empirical ratio of having any word after or before that specific word. Now, assuming these relations between all of the words exist, we have an object-dependent representation for each word based on its relation with all the other words. This would be a huge pre-specific representation of concrete objects. It is pre-specific because compared to the above-mentioned language models, in this mode of representation of words there is no specifically given semantic or syntactical property (e.g., synonym structures or grammatical rules) to the words. The whole network is constructed out of summation and division operations. However, as Markov says "many mathematicians apparently believe that going beyond the field of abstract reasoning into the sphere of effective calculations would be humiliating."[48]

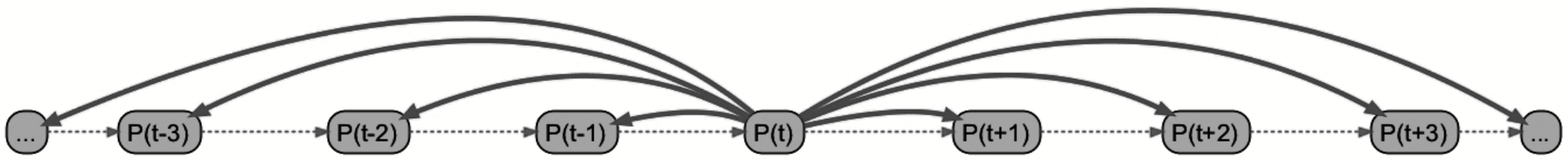

Figure 17. Language representation based on concrete universals: Markov's approach in constructing an empirical representation of a language's words based on a set of observed sentences.

Here again we have a self-referential setup, where concrete instances are implicitly represented by their relations to the other instances. As a result, if two particular words have the same function in that language, they will have similar relations with the other words. This was a big claim in 1906, when there wasn't even enough computing power to construct these relational networks.

As Claude Shannon later noted, even after almost forty years Markov's proposed modeling framework was not practically feasible, since it demands a large number of observations and relatively large computational power.[49] Nevertheless, as mentioned in section 7, the recent rapid growth in computation power has changed the situation dramatically and a similar approach, distributed representation, has attracted many researchers and practitioners.[50] Further, new applications of neural probabilistic models of language are becoming popular, while classical approaches in natural language processing are still struggling for a real world and scalable application.[51]

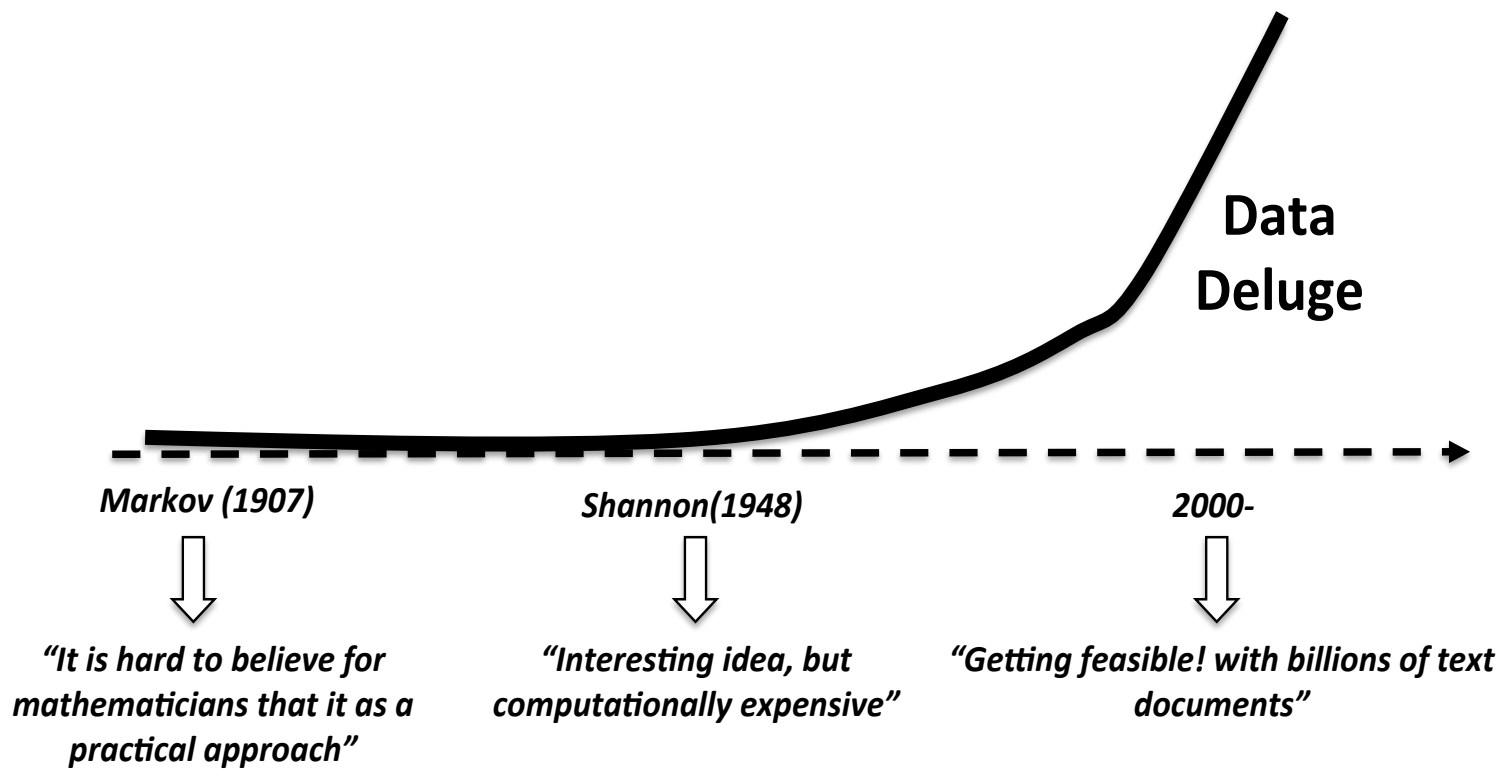

Figure 18. Distributed representation of linguistic models from Markov to the age of data deluge

The PageRank algorithm, used in Google searches, is another important application of Markov chains that follows the concept of representation of objects in a concrete level.[52] Around the year 2000, due to exponential growth in the number and diversity of webpages, the ranking of search results yielded by Internet search engines was becoming a critical issue. Prior to PageRank, most solutions sought to define a set of features for each webpage and then apply a scoring logic based on these features. In other words, the starting point for the ranking system was the act of defining a generic webpage, represented with a set of abstract universals. Consequently, every particular page would be a point in this parametric space of the generic webpage. Statistically, what happens is that if we increase the size of observations (i.e., the number webpages), $N_o$, the ratio of the number of parameters (dimensions), $N_p$, to the number of observations, $N_p/ N_o$, quickly gets close to zero. This means that the parametric space becomes full, and the discrimination of different webpages from each other becomes impossible. Therefore, considering the size and the diversity of webpages across the Internet, the process of defining features and assigning values to each page was a bottleneck and theoretically limited.

With the Markov-based procedure developed in PageRank, Google did not improve the classic approach, but rather changed the paradigm. They changed the basic assumption of centralized ranking and simply assumed that individuals know best which pages are related and important for them—better than any axiomatic or semantic order could know. They looked on a micro-scale at how individuals link important webpages to their webpages, and based on these live streams of data, they constructed, continuously and adaptively, a Markov chain, mapping how people likely surf the Internet and thereby they constructed a probabilistic network of connections within the pages. They defined the importance of a webpage as the result of the importance of the webpages connected to that web page, which is a self-referential equation with no externally imposed feature set. In terms of statistical representation, what happens is that this object-dependent representation is scalable to the size of the observations (webpages), since each concrete objects brings its representation with itself and acts as a new parameter for the representation of all the other pages (for simplicity's sake, assume a binary relation of a webpage to the other pages). This practical application conceptually is aligned with the definition of the Dedekind cut, where instead of representing an object directly, each object is represented as the negation of its complement.

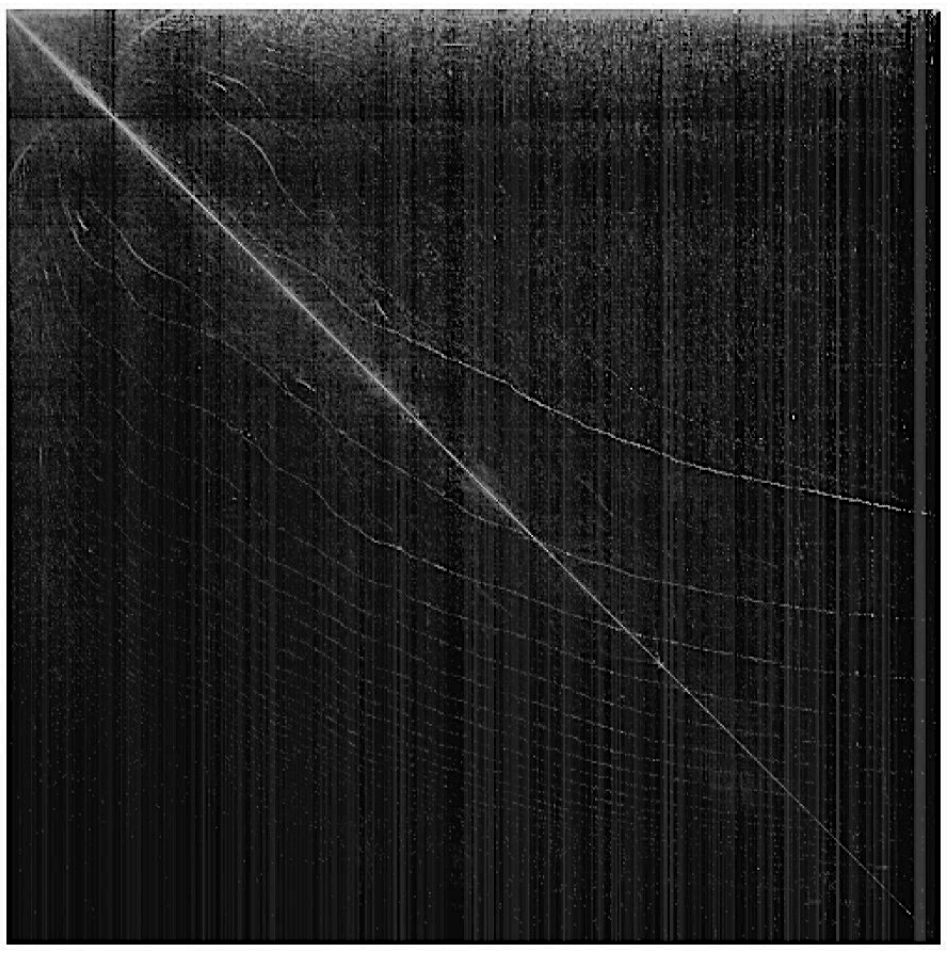

Figure 19. A part of Google matrix: I=HI. Everything is represented based on every other thing.

Fortunately, theories from linear algebra are available to solve this self-referential equation. The values of first Eigenvector of the constructed Markov matrix are used as the ranking of the pages. With the same methodology, it is possible to model similar problems in other fields. For example, a Markov chain based on available GPS trackings of cars can be used for modeling traffic dynamics in an urban street network,[53]

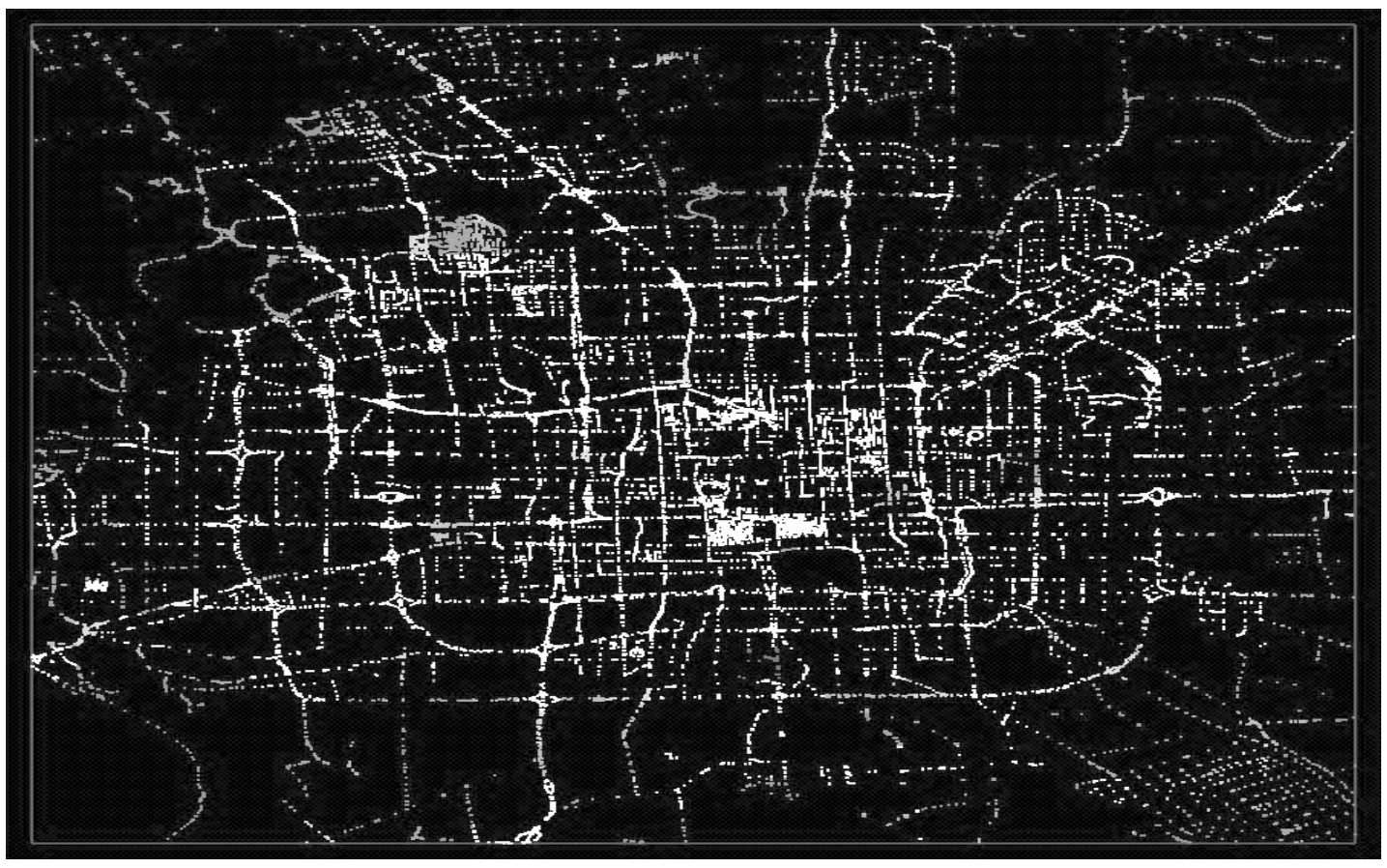

Figure 20. A Markov chain-based representation of traffic networks from sequence of GPS tracks of cars in Beijing

## 8.2. Self-Organizing Map

Following the same line of argumentation for the issue of representation in complex systems that we had for Markov chains, there is another powerful data-driven, pre-specific modeling method called the Self-Organizing Map (SOM).[54] As a well-known method in machine learning, the SOM has a very rich literature with a diverse set of applications.[55] According to the literature, the SOM is a generic methodology that has been applied in many classical modeling tasks such as the visualization of a high-dimensional space,[56] clustering and classification,[57] and prediction and function approximation.[58] During the past three decades there have been different extensions and modifications to the original algorithm that was introduced by Teuvo Kohonen in 1982. For example, one can compare the SOM with other clustering methods or with space transformation and feature extraction methods such as the Principal Component Analysis (PCA).[59] It is possible to explain and compare the SOM with vector quantization methods.[60] Further, it is possible to explain the SOM as a nonlinear function approximation method and to see it as a type of neural network methods and radial basis function.[61]

However, in this work I present two main aspects of the SOM in relation to the idea of pre-specific representation and in comparison with other modernist mathematical approaches, which are based on the notions of ideals and abstract universals.

### 8.2.1. No More External Dictionary and No More Generic Object

As discussed in section 5, in specific modeling the observations of any real phenomena are encoded in a generic object being represented by a set of given parameters. The underlying idea of pre-specific modeling is how to relax the modeling process from any specific and idealistic representation of the real phenomena—or, how not to depend on the generic object. For example, as mentioned before, in a Fourier transformation we assume that any dynamic behavior can be reconstructed and re-presented by a set of ideal cyclic forms. As figure 21 shows, an observed signal can be decomposed or can be approximated as a linear summation of some ideal waves. In other words, here we assume that there is a generic sinusoidal wave (as an ideal behavior) with a parametric setup, and by changing the parameters of this generic function there are different instances of waves. Finally, an observed signal can be represented as a summation of these ideal waves as follows:

$$s(t) = \frac{a_0}{2} + \sum_{n=1}^{\infty} [a_n cos(n\omega t) + b_n \sin(n\omega t)]$$

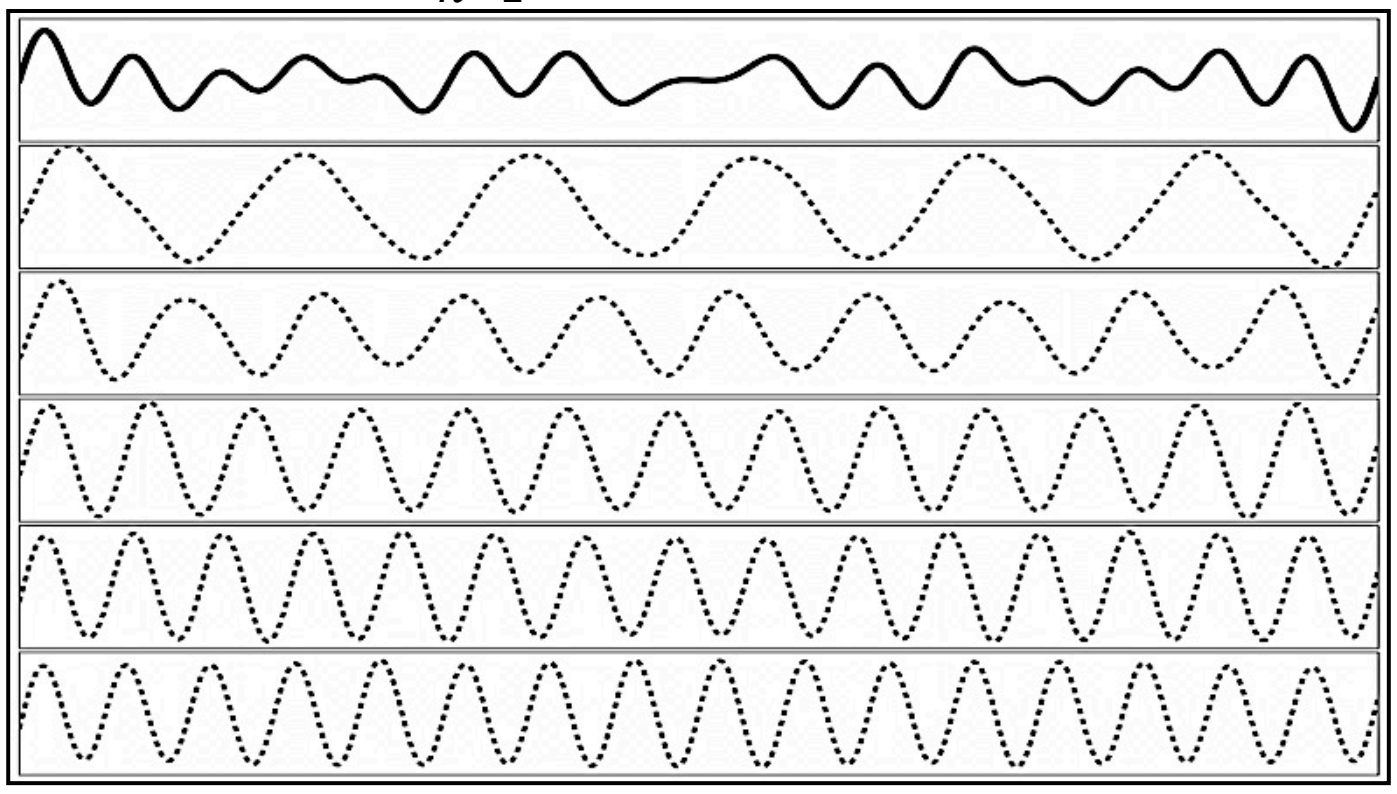

Figure 21. Reconstruction of an observed signal (top row) based on a parametric dictionary of ideal waves in Fourier decomposition

Therefore, the Fourier transformation is among the specific models that are based on abstract universals. As discussed, although powerful and useful in many classical engineering and scientific applications, this approach of idealized modeling has fundamental limits in dealing with complex (multifaceted) phenomena. The opposite idea or the complementary idea in pre-specific modeling is that based on the concept of concrete universals, it might be possible to establish a self-referential setup using concrete objects (i.e., the observations) to model real phenomena without any external representation or any external control. In the domain of machine learning, this is the underlying idea of unsupervised learning. Interestingly, the SOM corresponds with the idea of representation based on concrete universals. Comparing to Fourier decomposition, shown in figure 21, if we train a SOM with enough observations, we get a dictionary of potential dynamic forms collected via real observations (figure 22).

In assuming each of the prototypical forms in the trained SOM as a word or a letter in a language, a trained SOM can be used as pre-specific dictionary for the target phenomena. In other words, in terms of signal processing, assuming a fixed segmentation size, the observed signal can be translated into a set of numerical indexes (i.e., the index of the matching prototypes in the SOM network with each segment of the observation vector)—further, these indexes will be used for further steps of modeling. The main point is that unlike the case of Fourier decomposition, here there is no external axiomatic setup for the transformation of observations into codes; the whole encoding system provided by the SOM is based on concrete observations.

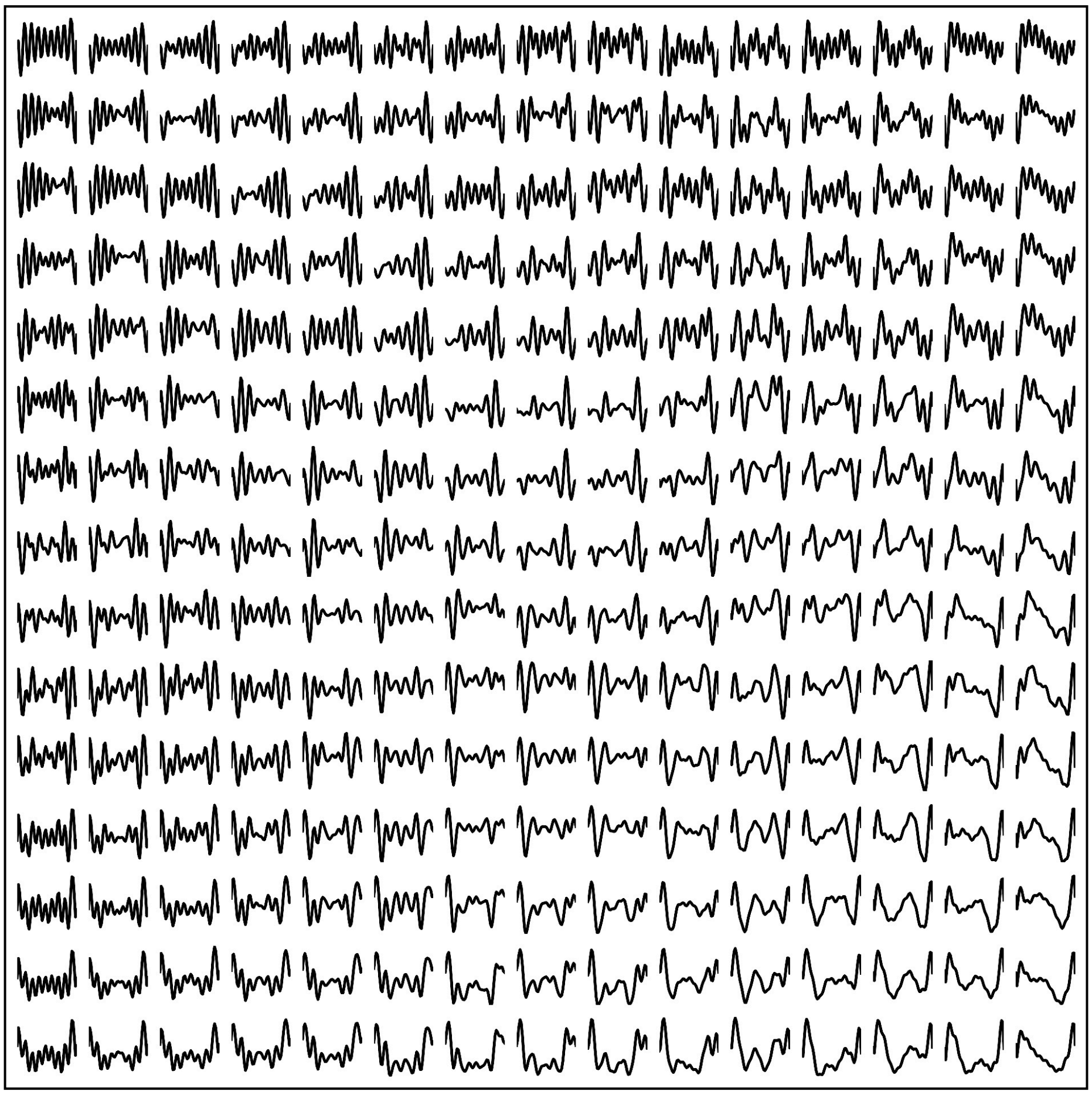

Figure 22. Self-organizing maps: constructing a pre-specific dictionary of dynamic forms from a large collection of observed signals.

Further, in a certain topology of SOM networks, the final indexes can be used to transform a multidimensional dynamic system into a one-dimensional symbolic dynamic system. In this case, the indexes of the SOM can be considered as "contextual numbers."[62]

In the field of computer vision and speech processing there has been a growing trend of methods that are based on the idea of representation that outperform many classical pattern recognition methods only in coexistence with a large amount of observations based on feature engineering. Classically, feature engineering means that in order to develop a pattern recognition model (i.e., in an image classification problem), one first needs to design a feature space to transform the images by that and then to develop a classification model on top of the engineered features. In the example of the Fourier analysis, the frequency and phase difference are the features. On the other hand, in this new category of modeling, some times called representation learning, there is no specific and separate feature-engineering task before fitting the classification or prediction method.[63] Among these algorithms, is the sparse coding algorithm, which has conceptual similarities with the SOM.[64] The principle idea of sparse coding is that if the original observations are n dimensional vectors, one can finds an over complete set of vectors (i.e., K vectors, where K>>n) to reconstruct the original observations with a linear and sparse combination of these K vectors. While it looks similar to methods such as PCA[65] or Independent Component Analysis (ICA),[66] sparse coding (similar to the SOM algorithm) does not produce a global transformation matrix. In PCA for example, all the n orthogonal basis vectors proportionally (according to their corresponding eigenvalues) contribute to the representation of all of the original observations, but in the SOM and sparse coding we have a kind of "distributed representation," in which each original observation is directly represented by a few specific prototypes (basis vectors). In other words, in the SOM each prototype is an object, which is not true for each principal component in PCA. They are from different worlds.

Further, this encoding approach can be applied in a hierarchical process. For example, in the case of image processing it can be applied to small patches of an image, where each patch will be indexed to a few codes and the next level (for example the whole image) will then be represented by new codes constructed on top of the previous codes. In fact, the output of one step is used as input for the next layer. Therefore, the whole image is analyzed by multilevel sparse codes. This simple idea of coding in an unsupervised approach has been applied in many practical applications and it has been claimed that it works better than the wavelet decomposition method.[67] I should note that the wavelets act similarly to the Fourier series, but they are more advanced, since there is no longer the assumption that the underlying ideal waves are stationary. Figure 23 shows an example of a sparse coding algorithm applied to image patches.

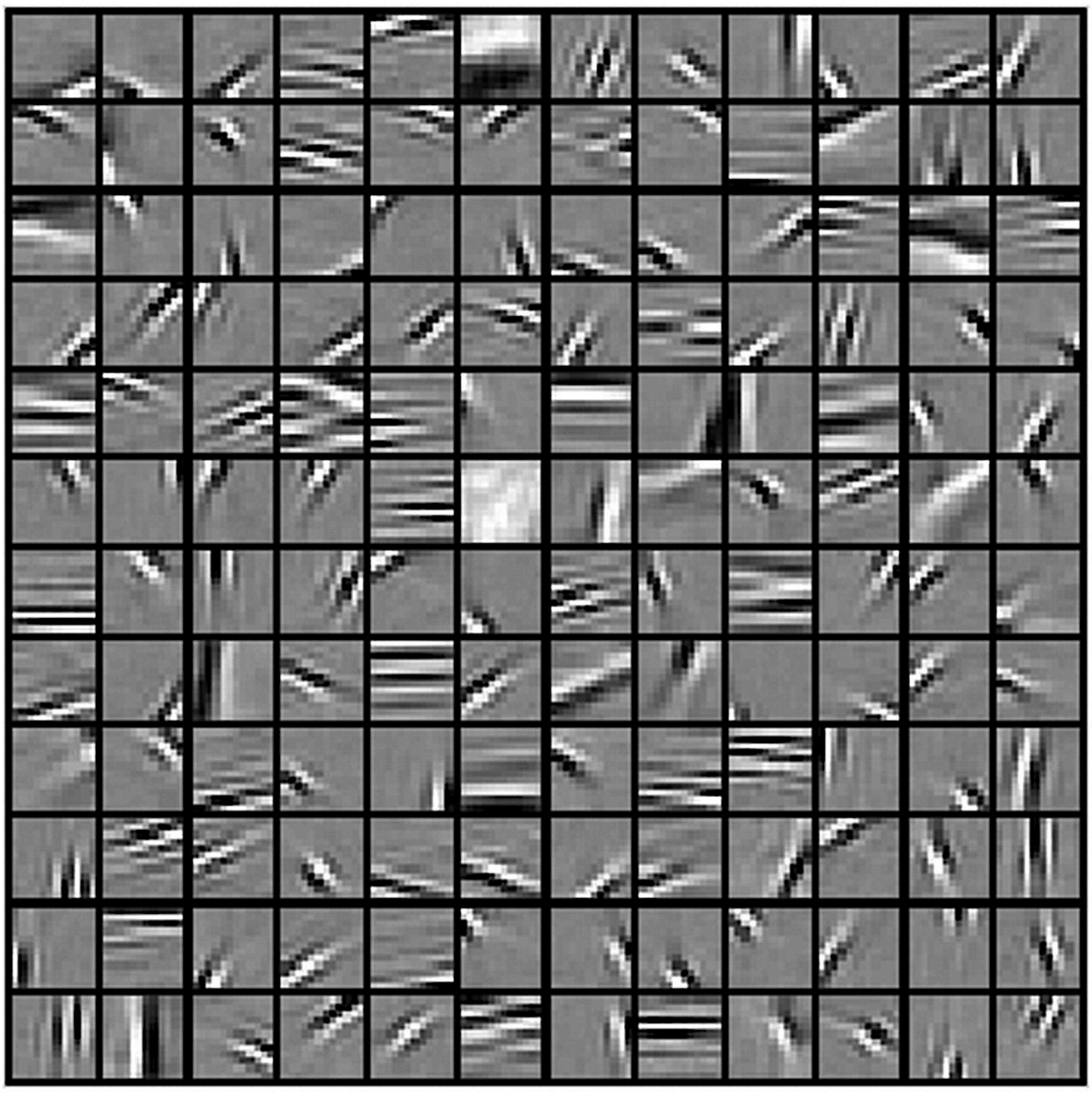

Figure 23. Sparse coding: learning data driven image dictionaries[68]

### 8.2.2. Computing with Indexes Beyond Ideal Curves

Another property of the SOM is its unique disposition for structural learning. Figure 24 shows the main difference between the SOM and a classical way of relation (function) modeling. In simple terms, the primary goal of relation modeling is to find the relation between two dimensions, based on a set of observations. In a classical way of modeling, one needs to fit a curve (a fixed structure) to a data set, while minimizing the deviations (errors) from the selected curve. In other words, the selected curve represents the logic that idealizes the observed data into a continuous relation. The SOM assumes that the logics (the argument that integrates cases) can be extracted from within the observed data—and it conserves all the logics (arguments) according to which it clusters the cases. What is optimized, in such modeling, is not how the data fits to logic, but the logic, which is being engendered, as much as possible from the data.

In this sense, in an analogy to a governance and decision-making system, we might say that the classical approach in curve fitting is a democratic setup, in which there is a global structure, tuned locally by the effect of individual votes. On the other hand, the SOM provides a social environment, in which each individual instance is not reduced but is kept active in its own individuality, while individuals can be unified into local clusters, if necessary.

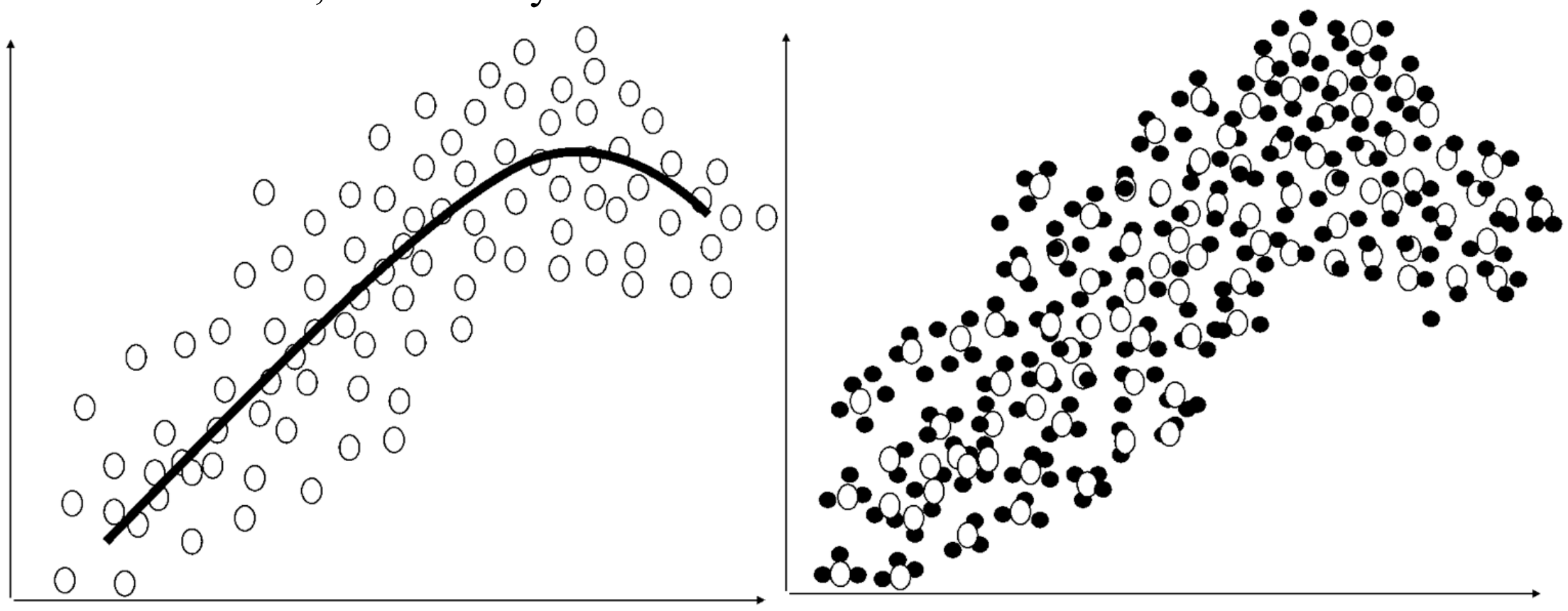

Figure 24. Computing with indexes beyond Ideal curves

Here again, the final model of the real phenomena is an abstraction of any potential specific model and it does not import any axiomatic or semantic specificity. Further, if the real environment is dynamic and evolving, and if we can assume the availability of dynamic data streams, then the SOM evolves along with the environment. In other words, pre-specific models are models in coexistence with data streams.

[7] Ibid.
[8] Max Black, *Models and Metaphors: Studies in Language and Philosophy* (Ithaca, NY: Cornell University Press, 1962).
[9] Ernan McMullin, “Galilean Idealization,” *Studies in History and Philosophy of Science Part A* 16, no. 3 (1985): 47–73.
[10] Mary B. Hesse, *Models and Analogies in Science* (London: Sheed and Ward, 1963).
[11] Ernan McMullin, “What do physical models tell us?” *Studies in Logic and the Foundations of Mathematics* 52 (1968): 385–96.
[12] Gonzalo Rodriguez-Pereyra, “What is the problem of universals?” *Mind* 109, no. 434 (2000): 255–73.
[13] McMullin, “Galilean Idealization”; Leszek Nowak, “Laws of Science, Theories, Measurement,” *Philosophy of Science* XXXIX (1972): 533–48; Michael Weisberg, “Three Kinds of Idealization,” *The Journal of Philosophy* (2007): 639–59.
[14] Weisberg, “Three Kinds of Idealization.”
[15] Mark Newman, *Networks: An Introduction* (Oxford: Oxford University Press, 2010).
[16] Luís M. A. Bettencourt et al., “Growth, Innovation, Scaling, and the Pace of Life in Cities,” *Proceedings of the National Academy of Sciences* 104, no. 17 (2007): 7301–7306.
[17] Michael Batty, “The Size, Scale, and Shape of Cities,” *Science* 319, no. 5864 (February 8, 2008): 769–71.
[18] David Grahame Shane, *Recombinant Urbanism: Conceptual Modeling in Architecture, Urban Design, and City Theory* (Chichester: Wiley, 2005).
[19] Tilmann Gneiting and Adrian E. Raftery, “Weather Forecasting with Ensemble Methods,” *Science* 310, no. 5746 (2005): 248–49.
[20] John Sterman, *Business Dynamics: Systems Thinking and Modeling for a Complex World* (New York: McGraw Hill, 2000).
[21] Waldo Tobler, “Cellular Geography,” in *Philosophy in Geography*, eds. Stephen Gale and Gunnar Olsson (Boston: D. Reidel, 1979), 379–86.
[22] Benoit Mandelbrot, *The Fractal Geometry of Nature* (New York: W. H. Freeman & Co., 1982).
[23] Matthew R. Evans et al., “Do simple models lead to generality in ecology?” *Trends in Ecology & Evolution* 28, no. 10 (2013): 578–83; Matthew R. Evans et al., “Data Availability and Model Complexity, Generality, and Utility: A Reply to Lonergan,” *Trends in Ecology & Evolution* 29, no. 6 (2014); Mike Lonergan, “Data Availability Constraints Model Complexity, Generality, and Utility: A Response to Evans et al.,” *Trends in Ecology & Evolution* 29, no. 6 (2014).
[24] Ellerman, D. P. (1988). Category theory and concrete universals. *Erkenntnis*, *28*(3), 409-429.
[25] Rodriguez-Pereyra, G. (2000). What is the Problem of Universals?. *Mind*, *109*(434), 255-273.
[26] Ellerman, “Category Theory and Concrete Universals,” PG 410
[27] Ibid., PG 410
[28] Ibid., PG 411
[29] David Ellerman, “On Concrete Universals: A Modern Treatment Using Category Theory” (May 10, 2014).., PG 6.
[30] Saunders Mac Lane and Ieke Moerdijk, *Sheaves in Geometry and Logic: A First Introduction to Topos Theory* (New York: Springer, 1992).
[31] Ellerman, D. (2014). On Concrete Universals: A Modern Treatment using Category Theory.
[32] Shillcock, R. (2014). The Concrete Universal and Cognitive Science. *Axiomathes*, *24*(1), 63-80.
[33] Panu Raatikainen, “Gödel's Incompleteness Theorems,” *The Stanford Encyclopedia of Philosophy* (Winter 2014 Edition), ed. Edward N. Zalta, http://plato.stanford.edu/archives/win2014/entries/goedel-incompleteness/.
[34] Ellerman, D. P. (1988). Category theory and concrete universals. *Erkenntnis*, *28*(3), 409-429.
[35] Wassermann, K. (2013). That Centre-Point Thing—The Theory Model in Model Theory. In *Applied Virtuality Book Series Printed Physics—Metalithikum I* (pp. 156-188). Springer Vienna.

[36] Richard Bellman, *Adaptive Control Processes: A Guided Tour* 4 (Princeton, NJ: Princeton University Press, 1961).
[37] Shillcock, R. (2014). The Concrete Universal and Cognitive Science. *Axiomathes*, *24*(1), 63-80.
[38] Bühlmann, V., & Wiedmer, M. (Eds.). (2008). *Pre-Specifics: Some Comparatistic Investigations on Research in Design and Art*. Zurich: JRP Ringier.
[39] Richard Dedekind, "The Nature and Meaning of Numbers," in *Essays on the Theory of Numbers*, trans. W. W. Beman (Chicago, IL: Open Court Publishing, 1901 [1898]).
[40] Ludger Hovestadt, Vera Buhlmann, "Eigen architecture", Ambra; Bilingual edition, 2014.
[41] Van Fraassen, B. C. (2010). Scientific representation: Paradoxes of perspective.
[42] Moosavi, V. (2015, April). Computational Urban Modeling: From Mainframes to Data Streams. In *AAAI Workshop: AI for Cities*.
[43] Lee Jr, D. B. (1973). Requiem for large-scale models. *Journal of the American Institute of Planners*, *39*(3), 163-178.
[44] Greenfield, A. (2010). *Everyware: The dawning age of ubiquitous computing*. New Riders.
[45] Hansmann, U., Merk, L., Nicklous, M. S., & Stober, T. (2003). *Pervasive computing: The mobile world*. Springer Science & Business Media.
[46] Chomsky, N. (2000). Linguistics and brain science. *Image, language, brain*, 13-28.
[47] Shillcock, R. (2014). The Concrete Universal and Cognitive Science. *Axiomathes*, *24*(1), 63-80.
[48] Basharin, G. P., Langville, A. N., & Naumov, V. A. (2004). The life and work of AA Markov. *Linear Algebra and its Applications*, *386*, 3-26.
[49] Shannon, C. E. (1948). A mathematical theory of communication, Part I, Part II. *Bell Syst. Tech. J.*, *27*, 623-656.
[50] Bengio, Y., Courville, A., & Vincent, P. (2013). Representation learning: A review and new perspectives. *IEEE transactions on pattern analysis and machine intelligence*, *35*(8), 1798-1828.
[51] Bengio, Y., Ducharme, R., Vincent, P., & Jauvin, C. (2003). A neural probabilistic language model. *Journal of machine learning research*, *3*(Feb), 1137-1155.
[52] Brin, S., & Page, L. (1998). The anatomy of a large-scale hypertexual web search engine. *Computer networks and ISDN systems*, *30*(1-7), 107-117.
[53] Moosavi, V., & Hovestadt, L. (2013, August). Modeling urban traffic dynamics in coexistence with urban data streams. In *Proceedings of the 2nd ACM SIGKDD International Workshop on Urban Computing* (p. 10). ACM.
[54] Kohonen, T. (1982). Self-organized formation of topologically correct feature maps. *Biological cybernetics*, *43*(1), 59-69.
[55] Kohonen, T. (2013). Essentials of the self-organizing map. *Neural networks*, *37*, 52-65.
[56] Vesanto, J. (1999). SOM-based data visualization methods. *Intelligent data analysis*, *3*(2), 111-126.
[57] Ultsch, A. (1993). Self-organizing neural networks for visualisation and classification. In *Information and classification* (pp. 307-313). Springer, Berlin, Heidelberg.
[58] Barreto, G. A., & Araujo, A. F. (2004). Identification and control of dynamical systems using the self-organizing map. *IEEE Transactions on Neural Networks*, *15*(5), 1244-1259.
[59] Yin, H. (2008). Learning nonlinear principal manifolds by self-organising maps. In *Principal manifolds for data visualization and dimension reduction*(pp. 68-95). Springer, Berlin, Heidelberg.
[60] Bishop, C. M., Svensén, M., & Williams, C. K. (1998). GTM: The generative topographic mapping. *Neural computation*, *10*(1), 215-234.
[61] Barreto, G. A., & Araujo, A. F. (2004). Identification and control of dynamical systems using the self-organizing map. *IEEE Transactions on Neural Networks*, *15*(5), 1244-1259.
[62] Moosavi, V. (2014). Computing With Contextual Numbers. *arXiv preprint arXiv:1408.0889*.
[63] Bengio, Y., Courville, A., & Vincent, P. (2013). Representation learning: A review and new perspectives. *IEEE transactions on pattern analysis and machine intelligence*, *35*(8), 1798-1828.
[64] Olshausen, B. A., & Field, D. J. (1996). Emergence of simple-cell receptive field properties by learning a sparse code for natural images. *Nature*, *381*(6583), 607.

[65] Pearson, K. (1901). LIII. On lines and planes of closest fit to systems of points in space. *The London, Edinburgh, and Dublin Philosophical Magazine and Journal of Science*, *2*(11), 559-572.
[66] Aapo, H., Juha, K., & Erkki, O. (2001). Independent component analysis. *Willey, New York*.
[67] Olshausen, B. A., & Field, D. J. (1996). Emergence of simple-cell receptive field properties by learning a sparse code for natural images. *Nature*, *381*(6583), 607.
[68] Hughes, J. M., Graham, D. J., & Rockmore, D. N. (2010). Quantification of artistic style through sparse coding analysis in the drawings of Pieter Bruegel the Elder. *Proceedings of the National Academy of Sciences*, *107*(4), 1279-1283.